\pdfoutput=1

\documentclass[11pt]{article}

\usepackage[]{acl}

\usepackage{times}
\usepackage{algorithm}
\usepackage{algpseudocode}
\usepackage{amsfonts}
\usepackage{amsmath}
\usepackage{amssymb}
\usepackage{booktabs}
\usepackage{caption}
\usepackage{graphicx}
\usepackage{latexsym}
\usepackage{paralist}
\usepackage{subcaption}
\usepackage{tabularx}
\usepackage{xcolor}
\usepackage{xspace}
\usepackage{enumitem}
\usepackage{float}
\usepackage{cleveref}

\usepackage[T1]{fontenc}

\usepackage[utf8]{inputenc}

\renewcommand{\paragraph}[1]{\medskip\noindent\textbf{#1}}

\newcommand{\newText}[1]{{\color{gray}#1}}
\renewcommand{\newText}[1]{#1}

\newcommand{\vulgar}{vulgar\xspace}
\newcommand{\vulgarity}{vulgarity\xspace}
\newcommand{\Vulgar}{Vulgar\xspace}

\newcommand{\OnI}{\textsc{OnI}\xspace}
\newcommand{\OI}{\textsc{OI}\xspace}
\newcommand{\Racist}{Anti-Black\xspace}
\newcommand{\racist}{anti-Black\xspace}
\newcommand{\AAE}{AAE\xspace}

\newcommand{\smallscale}{breadth-of-workers\xspace}
\newcommand{\Smallscale}{Breadth-of-Workers\xspace}
\newcommand{\largescale}{breadth-of-posts\xspace}
\newcommand{\Largescale}{Breadth-of-Posts\xspace}

\newcommand{\scale}[1]{\textsc{#1}\xspace}
\newcommand{\racistBeliefs}{\scale{RacistBeliefs}}
\newcommand{\freeSpeech}{\scale{FreeOffSpeech}}

\newcommand{\harmOfHateSpeech}{\scale{HarmOfHateSpeech}}
\newcommand{\empathy}{\scale{Empathy}}
\newcommand{\altruism}{\scale{Altruism}}
\newcommand{\traditionalism}{\scale{Traditionalism}}
\newcommand{\lingPurism}{\scale{LingPurism}}

\newcommand{\perspective}{\textsc{Perspective}\xspace}
\newcommand{\perspectiveAPI}{\textsc{PerspectiveAPI}\xspace}

\title{
    Annotators with Attitudes:\\How Annotator Beliefs And Identities Bias Toxic Language Detection 
}

\newcommand\uwcse{$^\heartsuit$}
\newcommand\uwpsych{$^\diamondsuit$}

\newcommand\aitwo{$^\spadesuit$}
\newcommand{\gatech}{$^\clubsuit$}
\newcommand{\aspace}{\hspace{.8em}}

\author{Maarten Sap\uwcse\aitwo\aspace 
    Swabha Swayamdipta\aitwo\aspace
    Laura Vianna\uwpsych\aspace
    \\
    \textbf{Xuhui Zhou}\gatech\aspace 
    \textbf{Yejin Choi}\uwcse\aitwo\aspace 
    \textbf{Noah A. Smith}\uwcse\aitwo\\\vspace{-0.3em}
    \small{\uwcse Paul G. Allen School of Computer Science, University of Washington, Seattle, WA, USA}\\ \vspace{-0.3em}
    \small{\aitwo Allen Institute for AI, Seattle, WA, USA}\\ \vspace{-0.3em}
    \small{\uwpsych Department of Psychology, University of Washington, Seattle, WA, USA}\\ 
    \small{\gatech Georgia Institute of Technology, Atlanta, GA, USA} \\
    \texttt{\{maartens,swabhas\}@allenai.org}
}

\begin{document}

\maketitle

    \begin{abstract}
    \textit{\color{red!55!black}\textbf{Warning}: this paper discusses and contains content that is offensive or upsetting.}

    The perceived toxicity of language can vary based on someone's identity and beliefs, but this variation is often ignored when collecting toxic language datasets, resulting in dataset and model biases. 
    We seek to understand the \textit{who}, \textit{why}, and \textit{what} behind biases in toxicity annotations.
    In two online studies with demographically and politically diverse participants, we investigate the effect of annotator identities (\textit{who}) and beliefs (\textit{why}), drawing from social psychology research
    about hate speech, free speech, racist beliefs, political leaning, and more.
    We disentangle \textit{what} is annotated as toxic by considering posts with three characteristics:
    \racist language, African American English (\AAE) dialect, and \vulgarity. 
    Our results show strong associations between annotator identity and beliefs and their ratings of toxicity.
    Notably, more conservative annotators and those who scored highly on our scale for racist beliefs were less likely to rate \racist language as toxic, but more likely to rate \AAE as toxic.
    We additionally present a case study illustrating how a popular toxicity detection system's ratings inherently reflect only specific beliefs and perspectives.
    Our findings call for contextualizing toxicity labels in social variables, which raises immense implications for toxic language annotation and detection.
    \end{abstract}

\section{Introduction}
\label{sec:intro}

\begin{figure*}[t]
    \centering
    \includegraphics[width=\textwidth]{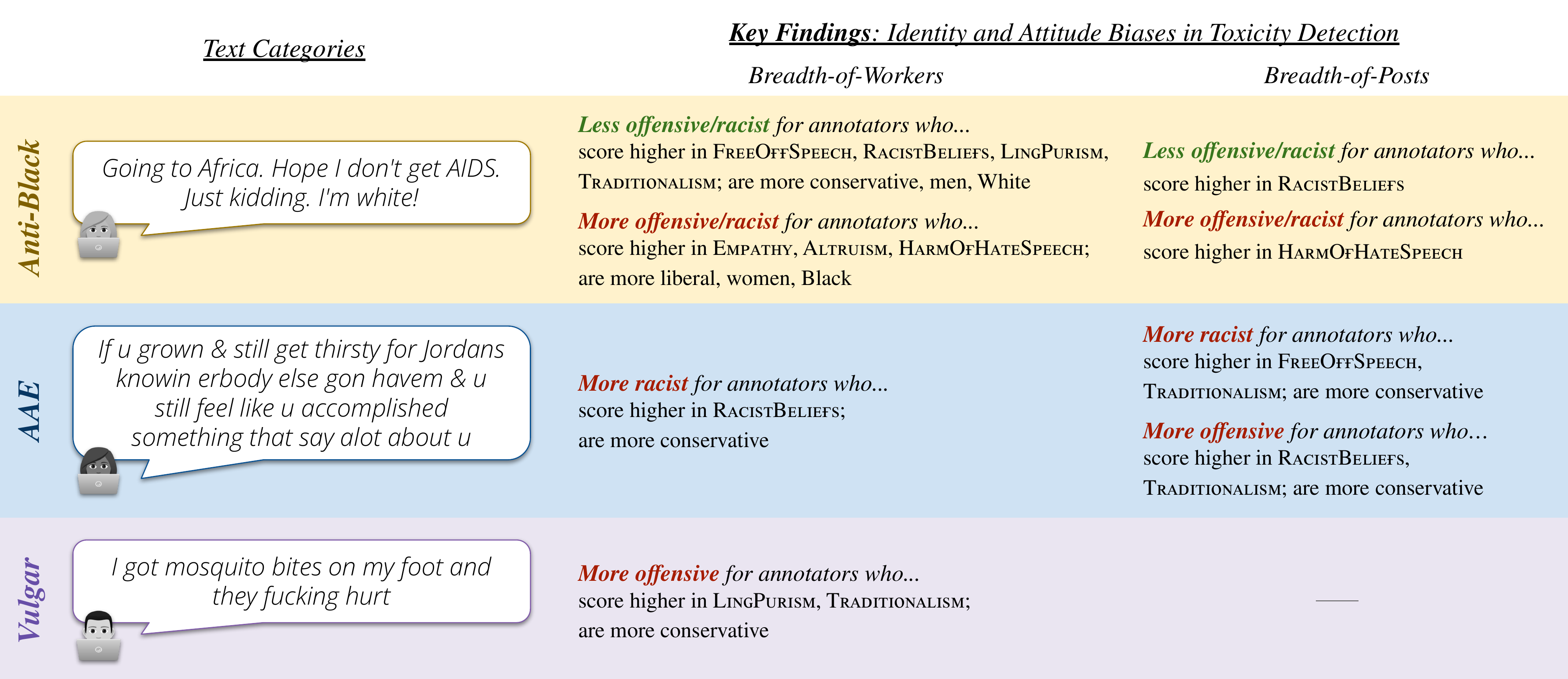}
    \caption{
    Annotator identities and attitudes can influence how they rate toxicity in text.  
    We summarize the key findings from our analyses of biases in toxicity (offensiveness or racism) ratings for three types of language: \racist content, African American English (\AAE), and \vulgar language.}
    \label{fig:introFig}
\end{figure*}

Determining whether a text is toxic (i.e., contains hate speech, abuse, or is offensive) is 
inherently a subjective task that requires a nuanced understanding of the pragmatic implications of language \cite{Fiske1993controlling,Croom2011slurs,waseem2021disembodied}.
Without this nuance, both humans and machines are prone to biased judgments, such as over-relying on seemingly toxic keywords \cite[e.g., expletives, swearwords;][]{dinan-etal-2019-build,han-tsvetkov-2020-fortifying} or backfiring against minorities \cite[i.a.]{Yasin2018black,Are2020-ej}.
For example, racial biases have been uncovered in toxic language detection where 
text written in African American English (\AAE) is falsely flagged as toxic \cite[][]{sap-etal-2019-risk,davidson-etal-2019-racial}.


The crux of the issue is that not all text is equally toxic for everyone \cite{waseem2016you,al-kuwatly-etal-2020-identifying}.
Yet, most previous research has treated this detection as a simple classification with one correct label, obtained by averaging judgments by a small set of human raters per post \cite{waseem-hovy-2016-hateful,wulczyn2017ex,davidson2017automated,founta2018large,Zampieri2019OLID}.
Such approaches ignore the variance in annotations \cite{Pavlick2019-tc,geva-etal-2019-modeling,Arhin2021-wa,akhtar2021whose} based on 
who the annotators are, and what their beliefs are.

In this work, we investigate the \textit{who}, \textit{why}, and \textit{what} behind biases\footnote{We use the term ``bias'' to denote both simple skews or variation in annotations (e.g., for variation in detecting \vulgar content as toxic) or representational harms \cite[e.g., \AAE being over-detected as toxic or \racist content being under-detected as toxic;][]{Barocas2017-bh,blodgett-etal-2020-language}.} 
in toxicity annotations, through online studies with demographically and politically diverse participants.
We measure the effects of annotator identities (\textit{who} annotates as toxic) and attitudes or beliefs (\textit{why} they annotate as toxic) on toxicity perceptions, 
through the lens of social psychology research on hate speech, free speech, racist beliefs, altruism, political leaning, and more.
We also analyze the effect of \textit{what} is being rated, by considering three text characteristics: \racist or racially prejudiced meaning, African American English (\AAE), and \vulgar words.

We seek to answer these questions via two online studies. 
In our \textbf{\smallscale} controlled study, we collect ratings of toxicity for a set of 15 hand curated posts from 641 annotators of different races, attitudes, and political leanings.
Then, in our \textbf{\largescale} study, we simulate a typical toxic language annotation setting by collecting toxicity ratings for $\sim$600 posts, from a smaller but diverse pool of 173 annotators.\footnote{Please contact the authors for the anonymized study data.}

Distilled in Figure \ref{fig:introFig}, 
our most salient results across both studies show that annotators scoring higher on our racist beliefs scale were less likely to rate \racist content as toxic (\S\ref{sec:racist-results}).
Additionally, annotators' conservatism scores were associated with higher ratings of toxicity for \AAE (\S\ref{sec:AAE-results}), and conservative and traditionalist attitude scores with rating \vulgar language as more toxic (\S\ref{sec:vulgar-results}).

We further provide a case study which shows that \perspectiveAPI, a popular toxicity detection system, mirrors ratings by annotators of certain attitudes and identities over others (\S\ref{sec:perspective-main}).
For instance, for \racist language, the system's scores better reflect ratings by annotators who 
score high on our scale for racist beliefs.
Our findings have immense implications for the design of toxic language annotation and automatic detection---we recommend contextualizing ratings in social variables and looking beyond aggregated discrete decisions (\S\ref{sec:discussion}).

\section{The \textit{Who}, \textit{Why}, and \textit{What} of Toxicity Annotations}
\label{sec:whowhatwhy}

We aim to investigate how annotators' ratings of the toxicity of text is influenced by their own identities (\textit{who they are}; \S\ref{ssec:who}), and their beliefs (\textit{why they consider something toxic}; \S\ref{ssec:why}) on specific categories of text (\textit{what they consider toxic}; \S\ref{sec:text_characteristics})---namely, text with \racist language, presence of African American English (\AAE), and presence of \vulgar or profane words.
To this end, we design two online studies (\S\ref{sec:study-designs}) 
and discuss \textit{who} find each of these text characteristics offensive and \textit{why} as separate research questions in Sections \S\ref{sec:racist-results}, \S\ref{sec:AAE-results}, and \S\ref{sec:vulgar-results}, respectively.

\subsection{Demographic Identities: \textit{Who} considers something as toxic?}
\label{ssec:who}

Prior work has extensively shown links between someone's gender, political leaning, and race affects how likely they are to perceive or notice harmful speech or racism \cite{Cowan2002-at,Norton2011-gw,Carter2015-lc,Prabhakaran2021-gx}.
Grounded in this prior literature, our study considers annotators' \textbf{race}, \textbf{gender}, and \textbf{political leaning}. 
Since perceptions of race and political attitudes vary vastly across the globe, we restrict our study to participants exclusively from the United States.

\subsection{Attitudes: \textit{Why} does someone consider something toxic?}
\label{ssec:why}
While some annotator toxicity ratings may highly correlate with demographic factors at face value \cite{Prabhakaran2021-gx,Jiang2021understanding}, we aim to go beyond demographics to investigate annotator \textit{beliefs} that explain these correlations.
Based on prior work in social psychology, political science, and sociolinguistics, we select seven attitude dimensions, which we operationalize via scales (in \textsc{small caps}), as described below.\footnote{We abstain from conclusions beyond our abstractions.}


\paragraph{Valuing the freedom of offensive speech (\freeSpeech):}
the belief that any speech, including offensive or hateful speech, should be unrestricted and free from censorship.
Recently, this belief has become associated with majority and conservative identities \cite{cole1996racist,Gillborn2009-rv,White2017-ad,Elers2020-ne}.
We use the scale by \citet{cowan2003empathy}; see Appendix \ref{app:free_speech}.

\paragraph{Perceiving the \harmOfHateSpeech:} 
the belief that hate speech or offensive language can be harmful for the targets of that speech \cite{Soral2018-wh,Nadal2018-iz}.
This belief is correlated with socially-progressive philosophies \cite[][see also \citeauthor{nelson2013marley}, \citeyear{nelson2013marley}]{Downs2012-lu}. 
We use the scale by \citet{cowan2003empathy}; see Appendix \ref{app:harm_hate_speech}.

\paragraph{Endorsement of \racistBeliefs:} 
the beliefs which deny the existence of racial inequality, or capture resentment towards racial minorities \cite{poteat2012modern}. 
We measure \racistBeliefs using items from the validated Modern Racism Scale \cite{mcconahay1986modern}; see Appendix \ref{app:racism}.

\paragraph{\traditionalism:}
the belief that one should follow established norms and traditions, and be respectful of elders, obedient, etc.
In the US, these beliefs are associated with generally conservative ideologies \cite{johnson2001social,knuckey2005new}.
We use an abridged version\footnote{This was done to reduce cognitive load on annotators.} of the \traditionalism scale \cite{bourchard2003genetic} that measures annotators' adherence to traditional values; see Appendix \ref{app:traditionalism}.

\paragraph{Language Purism (\lingPurism):} 
the belief that there is a ``correct'' way of using English \cite{jernudd1989politics}.
Typically, this belief also involves negative reactions to non-canonical ways of using language \cite{sapolsky2010rating,defrank2019language}.
We created and validated a four-item \lingPurism scale to measure this concept; see Appendix \ref{app:language_purism}.

\paragraph{\empathy:} 
one's tendency to see others' perspectives and feel others' feelings.
Research in social psychology has linked higher levels of empathy to the ability and willingness of recognizing and labeling hate speech \cite{cowan2003empathy}. 
We measure \empathy using an abbreviated Interpersonal Reactivity Index  \cite{pulos2004hierarchical};  see Appendix \ref{app:empathy}.

\paragraph{\altruism:}
one's attitude of selfless concern about others' well-being, which can move people to act when harm or injustice happens \cite{wagstaff1998equity,gavrilets2014solution,riar2020game}, including harms through hate speech \citep{Cowan2002-at}.
We gathered the items to measure \altruism with an adapted scale taken from \citet{steg2014significance}; see Appendix \ref{app:altruism}.

It is worth noting that some of the above attitudes, though not all, correlate with demographics very strongly.
Table \ref{tab:data-correls} in Appendix \ref{ssec:inter-variable-correls} details these correlations from our study.

\subsection{Text Characteristics: \textit{What} is considered offensive?}
\label{sec:text_characteristics}


Not all toxic text is toxic for the same reasons. 
We aim to understand how \textit{characteristics of text} can affect ratings of toxicity, in addition to annotator attitudes and identities.
Specifically, we consider three dimensions or categories of text, based on recent work on text characteristics that tend to be over- or under-detected as toxic \cite{dinan-etal-2019-build,sap-etal-2019-risk,han-tsvetkov-2020-fortifying,zhou2021challenges}:
\racist language, presence of African American Engligh (\AAE) dialect markers, and \vulgar language (e.g., swearwords, slurs).
\newText{We distinguish between two types of \vulgarity, following \citet{zhou2021challenges}: swearwords or explicit words that do not reference identities (\textbf{\underline{o}}ffensive, \textbf{\underline{n}}on-\textbf{\underline{i}}dentity referring; \OnI), and (reclaimed) slurs or other identity-referring vulgarity (\textbf{\underline{o}}ffensive \textbf{\underline{i}}dentity-referring; \OI). 
In our analyes, we focus on \OnI \vulgarity unless explicitly noted.}



\section{Data \& Study Design}
\label{sec:study-designs}


We design two online studies to study the effect of annotator identities and attitudes on their toxicity ratings on posts with different characteristics.
In either study, annotators are asked to rate how \textit{offensive} and how \textit{racist} they consider a post to be (see Appendix \ref{ssec:toxicity-questions-list} for the exact questions).\footnote{\newText{For both studies, we sought explicit consent from participants, paid participants above minimum wage, and obtained approval by our institution's ethics board (IRB).}}
\newText{We specifically focus on readers' perceptions or opinions, instead of imposing prescriptive definitions of toxicity or hate speech which previous work has shown still suffers from large annotator disagreement \cite{Ross2017measuring}.}
\newText{In the sections \S\ref{sec:racist-results}--\ref{sec:vulgar-results}, we report only (near-)significant associations; see Appendix \ref{sec:smallScale-more} and \ref{app:largeScale-results} for all results.}


\newcommand{\teaEmoji}{\includegraphics[height=1em]{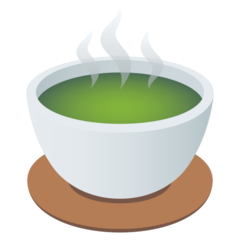}\xspace}
\newcommand{\handshakeEmoji}{\includegraphics[height=1em]{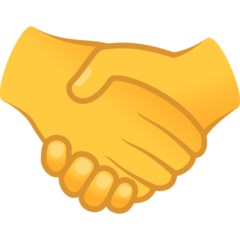}\xspace}
\newcommand{\smirkEmoji}{\includegraphics[height=1em]{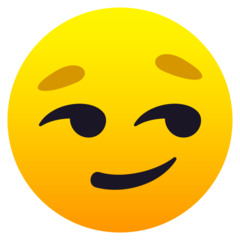}\xspace}

\newcommand{\tweet}[1]{{\fontfamily{pag}\selectfont\textit{#1} \vspace{.25em}}}

\begin{table}[t!]
\renewcommand{\arraystretch}{1.03}
\tiny
\begin{tabular}{@{}p{\columnwidth}@{}}

\multicolumn{1}{c}{\bf \centering \small \textit{\Smallscale study}\smallskip} \\ \toprule
\multicolumn{1}{c}{\bf \centering \Vulgar (\OnI), not \Racist, not \AAE} \\ \midrule
\tweet{I sort of fuckin love how close my gf is with my family like that she knows about my family jokes before I fuckin do} \\
\tweet{I'm hoping that Cersei gets her ass handed to her but I love Daenerys and Jon ;3; I don't want either of em losing xD} \\
\tweet{I got mosquito bites on my foot and they fucking hurt} \\
\tweet{Personally I think it would be fucking hilarious to put Franziska or Godot in but idk if they're important enough characters to make the cut} \\
\tweet{I just keep telling myself that shitty things happen to me but something really great is gonna happen bc romantic comedies have ruined me} \\ 
\toprule
\multicolumn{1}{c}{\bf \centering\AAE, not \Vulgar, not \Racist} \\ 
\midrule
\tweet{If u grown \& still get thirsty for Jordans knowin erbody else gon havem \& u still feel like u accomplished something that say alot about u} \\
\tweet{Just spent a whole 20k(imaginary money) on a home studio.. And used the last 3\$ 2 get some flowers n a gift card for YA BIHH.} \\
\tweet{When everything is going accordingly the enemy always try to come creep up on you but I aint havin it !! buhh byee} \\
\tweet{You can't slang no tea \teaEmoji cause he been putting me on bout everything \smirkEmoji. I could only wish dawg the best \handshakeEmoji} \\
\tweet{ain't no problem in cutting ppl off. save that space for somebody who'll benefit from u and vice versa} \\ 
\toprule
\multicolumn{1}{c}{\bf \Racist, not \Vulgar, not \AAE} \\ 
\midrule
\tweet{I don’t get government benefits just for being white} \\
\tweet{Where is white history month? Why can’t I have white pride day?} \\
\tweet{I don’t like black doctors, it’s just my preference} \\
\tweet{Anti-Black racism isn’t that big of a deal because people treated Italians like this in the 60’s and 70’s} \\
\tweet{Going to Africa. Hope I don't get AIDS. Just kidding. I'm white!} \\ \bottomrule
\end{tabular}
\caption{
All 15 posts for our \smallscale study.
}
\label{tab:study-tweets}
\end{table}

\subsection{\Smallscale Study}
\label{ssec:small-scale-design}
Our first study focuses on collecting toxicity ratings from a wide and diverse set of participants for a controlled set of posts.
Shown in Table~\ref{tab:study-tweets}, we \newText{hand} curated a set of 15 posts that belong exclusively to one text category (e.g., \vulgar but non-\AAE and non-\racist; see Appendix \ref{ssec:smallScale-data-selection} for more data selection and validation details).
To exclude confounding effects of offensive identity mentions \newText{(\OI; e.g., slurs) which could be both \vulgar and \racist (or sexist, homophobic, etc.)}, 
we only considered posts with \vulgar terms that are non-identity referring (\OnI; e.g., swearwords).

We ran our study on a 641 participants that were recruited using a pre-qualifier survey on Amazon Mechanical Turk (MTurk) to ensure racial and political diversity. 
Our final participant pool spanned various racial (13\% Black, 85\% White), political (29\% conservative, 59\% liberal), and gender identities (54\% women, 45\% men, 1\% non-binary).
Each participant gave each of the 15 posts toxicity ratings, after which they answered a series of questions about their attitudes and their identity.
We used three attention checks to ensure data quality.
For further details, please see Appendix \ref{sec:smallScale-moreSetup}.

In our subsequent analyses, we compute associations between the toxicity ratings and identities or attitudes by computing the effect sizes (Pearson $r$ correlation or Cohen's $d$) between the average toxicity rating of the posts in each category and annotator identities or attitude scores.

\subsection{\Largescale Study}
\label{ssec:large-scale-design}


\begin{table}[t!]
    \footnotesize
    \centering
    \begin{tabular}{@{}ccccc@{}}
        \multicolumn{5}{c}{\bf \centering \small \textit{\Largescale study}\smallskip} \\ \toprule
        \textit{cat.} & \Racist & \AAE & \Vulgar (\OnI) & \Vulgar (\OI) \\ \midrule
        \textit{count} & 113 & 270 & 196 & 217 \\
        \bottomrule
    \end{tabular}
    \caption{
    Counts for each text category for the 571 posts in our \largescale study.
    \OI: identity-referring \vulgarity, \OnI: non-identity referring \vulgarity; categories are explained in \S\ref{sec:text_characteristics}.
    Posts could belong to multiple categories (Figure \ref{fig:mturk-cat-counts} in Appendix \ref{app:largeScale-results}).
    }
    \label{tab:mturk-counts}
\end{table}

Our second study focuses on collecting ratings for a larger set of posts, but with fewer annotators per post to simulate a crowdsourced dataset on toxic language.
\newText{In contrast to the previous study, we consider \racist or \AAE posts that could also be \vulgar, and allow this \vulgarity to cover both potentially offensive identity references (\OI) as well as non-identity vulgar words (\OnI; see \S\ref{sec:text_characteristics}).
We do not consider posts that are \racist and \AAE, since the pragmatic toxicity implications of \racist meaning expressed in \AAE are very complex \cite[e.g., in-group language with self-deprecation, sarcasm, reclaimed slurs;][]{Greengross2008selfdeprecating,Croom2011slurs}, and are thus beyond the scope of this study.
}

\newText{We draw from two existing toxic language detection corpora to select 571 posts (Table~\ref{tab:mturk-counts}).
For \AAE and possibly \vulgar posts, we draw from \citet{founta2018large}, using an automatic \AAE detector by \citet{blodgett2016demographic}%
\footnote{\label{footnote:aae-detector-validity}\newText{The text-only \AAE detector by 
\citet{blodgett2016demographic} strongly correlates ($r$=.60) with more race-aware \AAE detectors \cite{sap-etal-2019-risk}.}} 
and the \vulgarity word list from \citet{zhou2021challenges} for detecting \OI and \OnI terms.
For \racist and possibly \vulgar posts, we select posts annotated as \racist in \citet{vidgen-etal-2021-learning}, using the same method by \citet{zhou2021challenges} for detecting \vulgar terms.}
See 
Appendix \ref{ssec:largeScale-dataSelection} for more data selection details.

As with the previous study, we ran our annotation study on 173 participants recruited through a pre-qualifier survey.
Our annotators varied racially (20\% Black, 76\% White), politically (30\% conservative, 54\% liberal), and in gender (45\% women, 53\% men, $<$2\% non-binary).
Each post was annotated by 6 participants from various racial and political identities.\footnote{For each post, we collected toxicity ratings from two white conservative workers, two from white liberal workers, and two from Black workers.}
Additionally, we asked participants one-item versions of our attitude scales, using the question from each scale that correlated best with toxicity in our \smallscale study as explained in Appendix \ref{ssec:largeScale-itemSelection}.
See Appendix \ref{sec:largeScale-moreSetup} for more study design details.

\newText{In our analyses, we examine toxicity of \racist and potentially \vulgar posts (\S\ref{ssec:racist-largescale-results}) and of \AAE and potentially \vulgar posts (\S\ref{ssec:aae-largescale-results}), but not of \vulgar posts separately, due to confounding effects of the \AAE or \racist characteristics that those posts could have.}
Additionally, unlike the \smallscale study, here each annotator could rate a varying number of posts. 
Thus, we compute associations between toxicity ratings and identities or attitudes using a linear mixed effects model\footnote{Using the statsmodels implementation: \url{https://www.statsmodels.org/stable/generated/statsmodels.formula.api.mixedlm.html}} 
with random effects for each participant. 
\section{Who finds \textbf{\racist} posts toxic, and why?}
\label{sec:racist-results}

\begin{table}
    \centering
    \includegraphics[width=\columnwidth,clip,trim=0 2em 0 0]{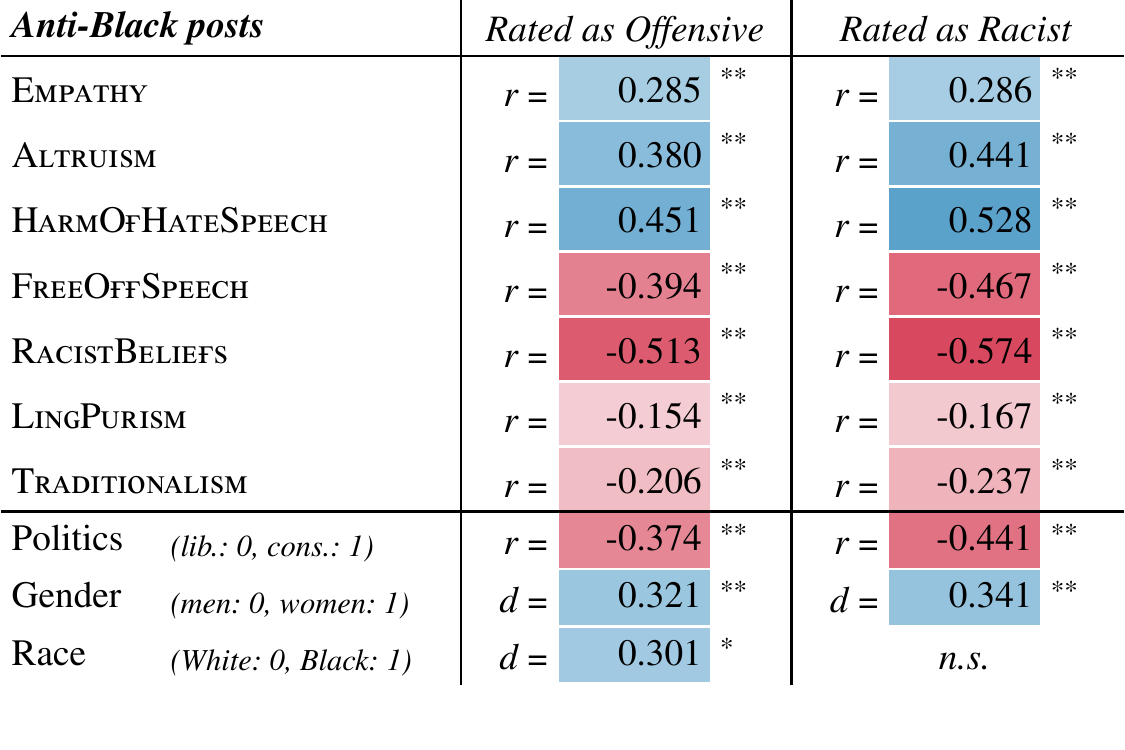}
    \caption{Associations between annotator variables and ratings of offensiveness and racism for the \textit{\racist} posts in the \textit{\smallscale} study. 
    We use the Holm correction for multiple comparisons for non-hypothesized associations and only present significant Pearson $r$ or Cohen's $d$ effect sizes ($^*$: $p$ < 0.05, $^{**}$: $p$ < 0.001; \textit{n.s.}: not significant).}
    \label{tab:smallScale-Racist}
\end{table}

\Racist language denotes racially prejudiced or racist content---subtle \cite{breitfeller-etal-2019-finding} or overt---which is often a desired target for toxic language detection research \cite{waseem2016you,vidgen-etal-2021-learning}. 
Based on prior work on linking conservative ideologies, endorsement of unrestricted speech, and racial prejudice with reduced likelihood to accept the term ``hate speech'' \cite{duckitt2003impact,White2017-ad,Roussos2018-sf,Elers2020-ne}, we hypothesize that conservative annotators and those who score highly on the \racistBeliefs or \freeSpeech scales will rate \racist tweets as less toxic, and vice-versa.
Conversely, based on findings by \citet{cowan2003empathy}, we hypothesize that annotators with high \harmOfHateSpeech scores will rate \racist tweets are more toxic.



\subsection{\Smallscale Results}

As shown in Table~\ref{tab:smallScale-Racist}, we found several associations between annotator beliefs and toxicity ratings for \racist posts, confirming our hypotheses.
The three most salient 
associations with \textit{lower racism} ratings were annotators who scored higher in \racistBeliefs, \freeSpeech, and those who leaned conservative.
We find similar trends for offensiveness ratings.

Conversely, we found that participants who scored higher in \harmOfHateSpeech were much more likely to rate \racist posts as \textit{more offensive}, and \textit{more racist}.
Finally, though both white and Black annotators rated these posts very high in offensiveness (with means $\mu_\text{Black}$ = 3.85 and $\mu_\text{white}$ = 3.59 out of 5), our results show that Black participants were slightly more likely than white participants to rate them as offensive.

Our exploratory analyses unearthed other significant associations: negative correlations with \lingPurism, 
\traditionalism, 
and gender (male), and positive correlations with high \empathy, \altruism, and gender (female). 

\subsection{\Largescale Results}
\label{ssec:racist-largescale-results}
\begin{table}[t]
    \centering
    \includegraphics[width=.8\columnwidth,clip,trim=0 13em 0 0]{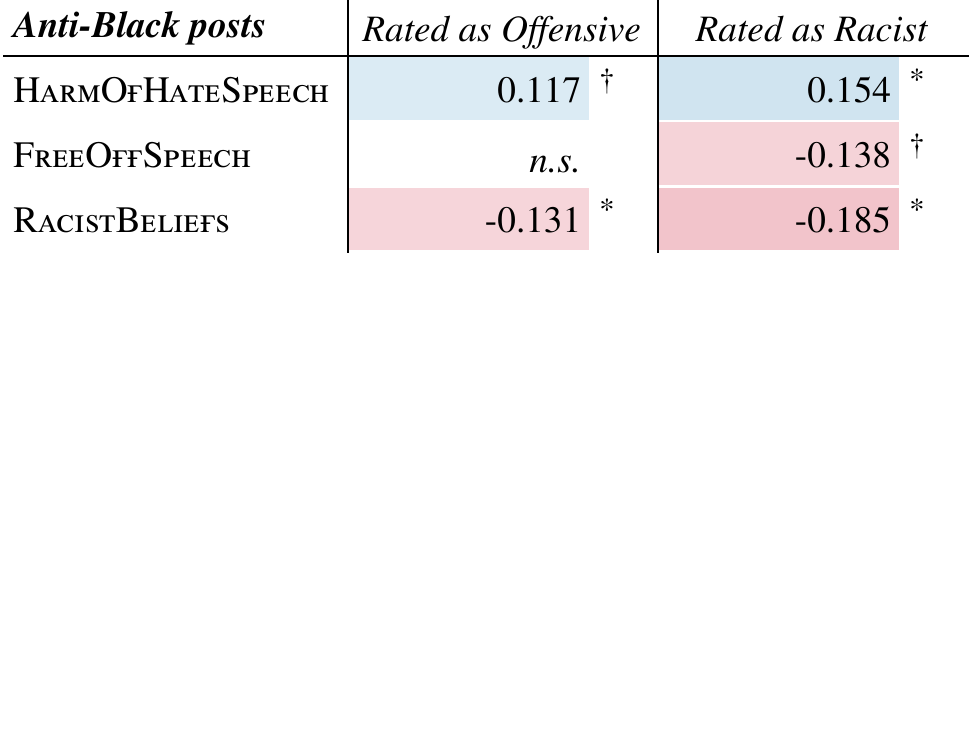}
    \caption{Associations for \textit{\racist} (and potentially also \vulgar) posts from the \textit{\largescale} study, shown as the $\beta$ coefficients from a mixed effects model with a random effect for each annotator ($^\dagger$: $p$ < 0.075, $^*$: $p$ < 0.05, $^{**}$: $p$ < 0.001; Holm-corrected for multiple comparisons;
    \textit{n.s.}: not significant).}
    \label{tab:largeScale-overlapping-Racist}
\end{table}

Table~\ref{tab:largeScale-overlapping-Racist}  shows similar results as in the \smallscale analyses, despite the posts now potentially containing \vulgarity.
Specifically, we find that annotators who scored higher in \racistBeliefs rated \racist posts as \textit{less} offensive, 
whereas those who scored higher in \harmOfHateSpeech rated them as \textit{more} offensive. 
Ratings of racism showed similar effects\newText{, along with a near-significant association between higher \freeSpeech scores and lower ratings of racism for \racist posts.}

\subsection{Perceived Toxicity of Anti-Black Language}
\label{ssec:discussion-Racist}

Overall, our results from both studies corroborate previous findings that studied 
associations between attitudes toward hate speech and gender and racial identities, specifically that conservatives, white people, and men tend to value free speech more, and that liberals, women, and non-white people perceive the harm of hate speech more \cite{cowan2003empathy,Downs2012-lu}.
Our results also support the finding that those who hold generally conservative ideologies tend to be more accepting towards \racist or racially prejudiced content \cite{Goldstein2017-mx,Lucks2020-ri,Schaffner2020-jv}.

In the context of toxicity annotation and detection, our findings highlight the need to consider the attitudes of annotators towards free speech, racism, and their beliefs on the harms of hate speech, for an accurate estimation of \racist language as toxic, offensive, or racist \cite[e.g., by actively taking into consideration annotator ideologies;][]{waseem2016you,vidgen-etal-2021-learning}.
This can be especially important given that hateful content very often targets marginalized groups and racial minorities \cite{silva2016analyzing,sap2020socialbiasframes}, and can catalyze violence against them \cite{OKeeffe2011impact,Cleland2014racism}.



\section{Who finds \textbf{\AAE} posts toxic, and why?}
\label{sec:AAE-results}

\begin{table}[t]
    \centering
    \includegraphics[width=.75\columnwidth,clip,trim=0 13em 0 0]{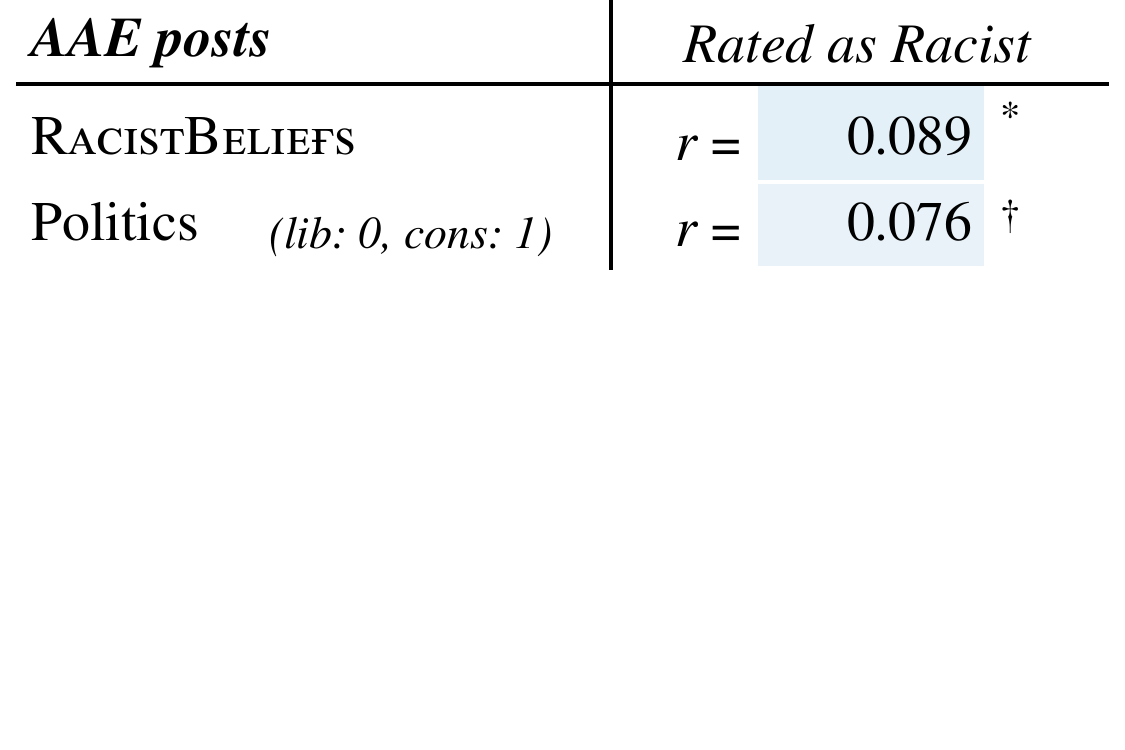}
    \caption{Associations between ratings of racism and annotator variables, for the \textit{\AAE} posts from the \textit{\smallscale} study. 
    As with the previous results, we correct for multiple comparisons for non-hypothesized associations and only show significant results ($^\dagger$: $p$ < 0.075, $^*$: $p$ < 0.05). 
    }
    \label{tab:smallScale-AAE}
\end{table}

African American English (\AAE) is a set of well-studied varieties or dialects of U.S. English, common among, but not limited to, African-American or Black speakers \cite{Green2002aae,edwards2004african}.
This category has been shown to be considered ``worse'' English by non-AAE speakers \citep{Hilliard1997-vx,Blake2003-ct,Champion2012-nv,Beneke2015-to,Rosa2017-ec}, and is often mistaken as obscene or toxic by humans and AI models \cite{spears1998african,sap-etal-2019-risk}, particularly due to dialect-specific lexical markers (e.g., words, suffixes).

Based on prior work that correlates racial prejudice with negative attitudes towards \AAE \cite{gaither2015sounding,Rosa2019soundingLikeRace}, we hypothesize that annotators who are white and who score high in \racistBeliefs will rate \AAE posts as more toxic.
Additionally, since \AAE can be considered non-canonical English \cite{sapolsky2010rating,defrank2019language}, we hypothesize that annotators who are more conservative and who score higher in \traditionalism and \lingPurism will rate \AAE posts with higher toxicity.



\subsection{\Smallscale Results}
\label{ssec:aae-smallscale-results}

Table~\ref{tab:smallScale-AAE} shows significant associations between annotator identities and beliefs and their ratings of toxicity of \AAE posts.
Partially confirming our hypothesis, we found that ratings of racism had somewhat significant correlations with annotators' conservative political leaning, 
and their scores on our \racistBeliefs scale. 
However, contrary to our expectations, we found that white and Black annotators did not differ in how offensive they rated \AAE tweets ($d$ = 0.14, $p$ > 0.1).
We found no additional hypothesized or exploratory associations for racism ratings, and no significant associations for offensiveness ratings. 

\subsection{\Largescale Results}
\label{ssec:aae-largescale-results}

\begin{table}[t]
    \centering
    \includegraphics[width=\columnwidth,clip,trim=0 11em 0 0]{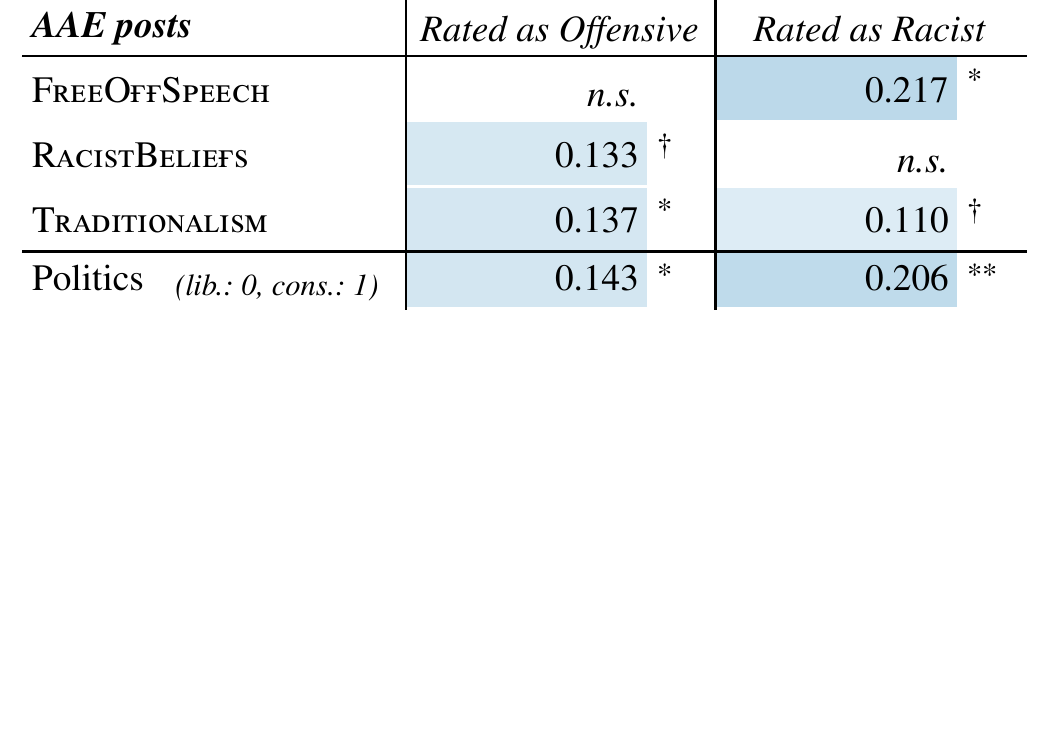}
    \caption{Associations between \textit{\AAE} (and potentially also \vulgar) post ratings from the \textit{\largescale} study and annotator variables, shown as the $\beta$ coefficients from a mixed effects model with a random effect for each annotator.
    We only show significant results ($^\dagger$: $p$ < 0.075, $^*$: $p$ < 0.05, $^{**}$: $p$ < 0.001; Holm-corrected for multiple comparisons; \textit{n.s.}: not significant).
    }
    \label{tab:largeScale-overlapping-AAE}
\end{table}
Shown in Table~\ref{tab:largeScale-overlapping-AAE}, our results for \AAE and potentially \vulgar \largescale study show higher offensiveness ratings from conservative raters, and those who scored higher in \traditionalism and, 
\newText{almost significantly,} \racistBeliefs.
We also find that conservative annotators and those who scored higher in \freeSpeech \newText{(and near-significantly, \traditionalism)} 
rated \AAE posts as more racist.

As an additional investigation, we measure whether attitudes or identities affects toxicity ratings of \AAE posts that contain the word ``\textit{n*gga},'' a (reclaimed) slur that has very different pragmatic interpretations depending on speaker and listener identity \cite{Croom2011slurs}.
Here, we find that raters who are more conservative tended to score those posts as significantly more racist ($\beta=0.465, p=0.003$; corrected for multiple comparisons).

\subsection{Perceived Toxicity of \AAE}
\label{ssec:discussion-AAE}

Our findings suggest that annotators perceive that \AAE posts are associated with the Black racial identity \cite{Rosa2019soundingLikeRace}, which could cause those who score highly on the \racistBeliefs scale to annotate them as racist, potentially as a form of colorblind racism \cite[e.g., where simply mentioning race is considered racist;][]{bonilla2006colorblindRacism}.
Moreover, specific markers of \AAE could have been perceived as obscene by non-\AAE speakers \cite{spears1998african}, even though some of these might be reclaimed slurs \cite[e.g., ``n*gga'';][]{Croom2011slurs,Galinsky2013-rw}.
Contrary to expectations, annotators' own racial identity did not affect their ratings of \AAE posts in our studies. 
\newText{Future work should investigate this phenomenon further, in light of the variation in perceptions of \AAE within the Black community \cite{Rahman2008-lu,Johnson2022-mp}, and the increased acceptance and usage of \AAE by non-Black people in social media \cite[][]{Ilbury2020-mb,Ueland2020-pt}.}

These findings shed some light on the racial biases found in hate speech detection \cite{davidson-etal-2019-racial,sap-etal-2019-risk}, partially explaining why \AAE is perceived as toxic.
Based on our results, 
future work in toxic language detection should account for this over-estimation of \AAE as racist.
For example, annotators could explicitly include speakers of \AAE, or those who understand that \AAE or its lexical markers are not inherently toxic, or are primed to do so \cite{sap-etal-2019-risk}.
Avoiding an incorrect estimation of \AAE as toxic is crucial to avoid upholding racio-linguistic hierarchies and thus representational harms against \AAE speakers \cite{Rosa2019soundingLikeRace,blodgett-etal-2020-language}.
\section{Who finds \textbf{\vulgar} posts toxic, and why?}
\label{sec:vulgar-results}

Vulgarity can correspond to non-identity referring swearwords (e.g., \textit{f*ck}, \textit{sh*t}; denoted as \OnI) or identity-referring slurs (e.g., \textit{b*tch}, \textit{n*gga}; denoted as \OI).
Both types of vulgarity can be mistaken for toxic despite also having non-hateful usages \cite[e.g., to indicate emotion or social belonging;][]{Croom2011slurs,Dynel2012Swearing,Galinsky2013-rw}.


\begin{table}[t]
    \centering
    \includegraphics[width=.75\columnwidth,clip,trim=0 10em 0 0]{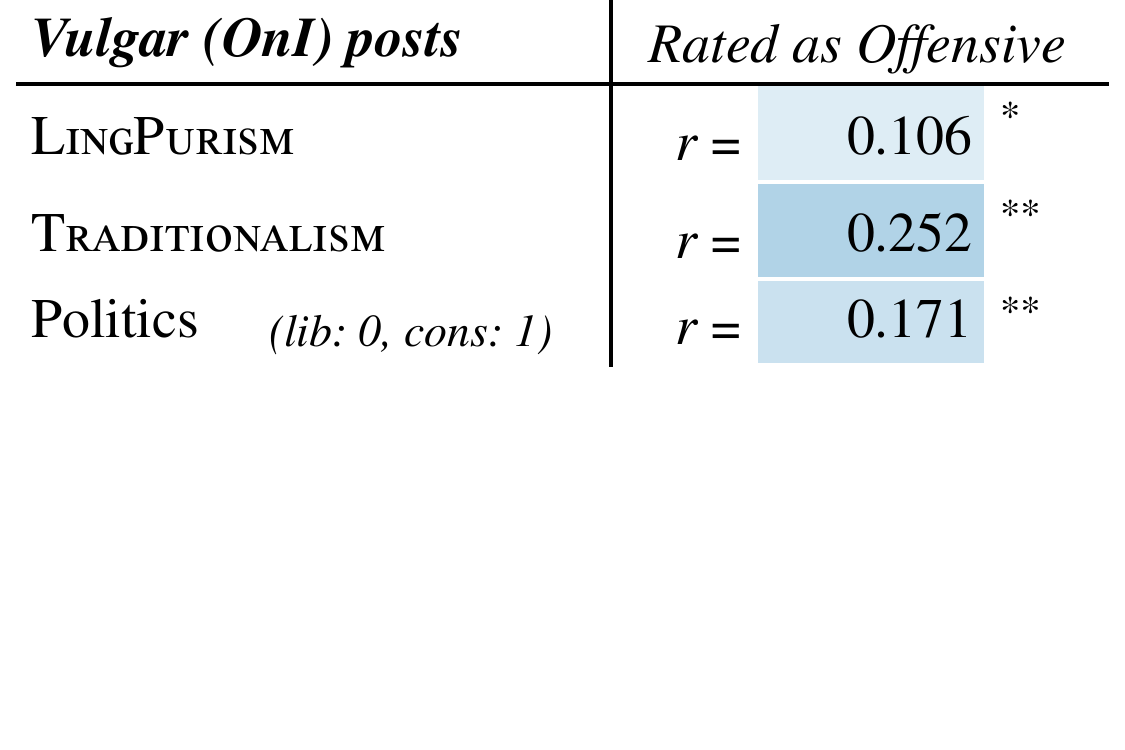}
    \caption{Associations between toxicity ratings and annotator variables for the \textit{\vulgar} posts from the \textit{\smallscale} study. 
    We correct for multiple comparisons for non-hypothesized associations and only show significant results ($^*$: $p$ < 0.05, $^{**}$: $p$ < 0.001). 
    }
    \label{tab:smallScale-Vulgar}
\end{table}

Given that vulgarity can be considered non-canonical or impolite language \cite{jay2008pragmatics,sapolsky2010rating,defrank2019language}, we hypothesize that annotators who score high on \lingPurism, \traditionalism, and who are more conservative will rate \vulgar posts as more offensive.
\ifdefined\keepLargeScaleVulgar 
    Importantly, we focus on the exclusively \vulgar (\OnI) posts from our \smallscale study, and consider both \vulgar (\OI and \OnI) in the \largescale study.
\else
    Importantly, here,  we focus on the posts that are exclusively \vulgar (\OnI) from only our \smallscale study, to avoid confounding effects of \vulgar posts with \racist meaning or in \AAE (both of those cases were analyzed in \S\ref{ssec:racist-largescale-results} and \S\ref{ssec:aae-largescale-results}).
    We refer the reader to Appendix \ref{app:largeScale-results} for the results on \vulgar posts in the \largescale study.
\fi



\subsection{\Smallscale Results}

Confirming our hypotheses, we found that offensiveness ratings of \vulgar (\OnI) posts indeed correlated with annotators' \traditionalism and \lingPurism scores, 
and conservative political leaning (Table~\ref{tab:smallScale-Vulgar}). 
We found no associations between attitudes and racism ratings for \vulgar posts.

\ifdefined\keepLargeScaleVulgar 

    \subsection{\Largescale results}
    \maarten{I almost feel like we shouldn't discuss the results for the \Largescale study for vulgar posts, because vulgar has high overlap with racist or AAE in this case, and results go in different directions?}
    \begin{table}[t]
        \centering
        \includegraphics[width=\columnwidth,clip,trim=0 7em 0 0]{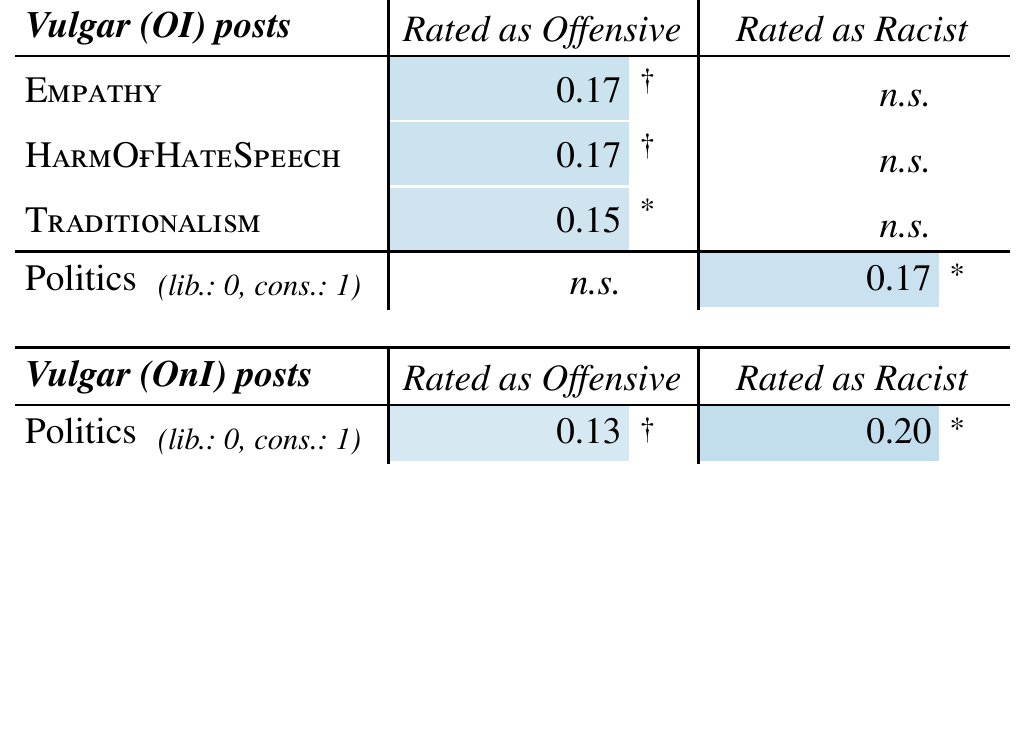}
        \caption{Associations on the posts that are \textit{\vulgar} (and potentially also \racist or \AAE) from the \textit{\largescale} study, shown as the $\beta$ coefficients from a mixed effects model with a random effect for each annotator.
        We only show significant results ($^\dagger$: $p<0.075$, $^*$: $p<0.05$, $^{**}$: $p<0.001$; Holm-corrected for multiple comparisons).}
        \label{tab:largeScale-overlapping-Vulgar}
    \end{table}
    
    Here, we find that annotators were more likely to rate \vulgar-\OI posts (e.g., with slurs) as offensive if they scored higher in \traditionalism, \empathy, and \harmOfHateSpeech. 
    Additionally, conservatives were more likely to rate these posts as racist \maarten{This is a finding that's weird}.
    Similarly, we find that conservatives were more likely to rate \vulgar-\OnI posts as more offensive and racist. 

\fi

\subsection{Perceived Toxicity of Vulgar Language}
\label{ssec:discussion-vulgar}

Our findings corroborate prior work showing how adherence to societal traditional values is often opposed to the acceptability of \vulgar language \cite{sapolsky2010rating}.
Traditional values and conservative beliefs have been connected to finding \vulgar language as a direct challenging the moral order \cite{Jay2018-yt,Sterling2020-cq,Muddiman2021-zz}.
Our results suggest that vulgarity is a very specific form of offensiveness that deserves special attention.
\ifdefined\keepLargeScaleVulgar 
    Given our findings in the \largescale study, future work might further differentiate vulgarity in language more explicitly, such as treating those that mention identities and those that do not, \textit{independently} \cite{zhou2021challenges}, and consider the individual vs. additive effects of \vulgarity used in text with offensive meaning.
\else 
    Specifically, future work might consider studying the specific toxicity of individual identity-referring \vulgar (\OI) words, which can carry prejudiced meaning as well (e.g., slurs such as ``\textit{n*gg*r}'').
\fi
Moreover, annotators across different levels of traditionalism could be considered when collecting ratings of vulgarity, especially since perceptions might vary with generational and cultural norms \cite{Dynel2012Swearing}.



\section{Toxicity Detection System Case Study: \perspectiveAPI}
\label{sec:perspective-main}

\definecolor{highBlue}{HTML}{2166ac}
\definecolor{lowRed}{HTML}{b2182b}
\newcommand{\rHigh}{r_{\text{high}}}
\newcommand{\rLow}{r_{\text{low}}}

Our previous findings indicated that there is strong potential for annotator identities and beliefs to affect their toxicity ratings.
We are additionally interested in how this influences the behavior of toxicity detection models trained on annotated data.
We present a brief case study to answer this question with the \perspectiveAPI,\footnote{\url{www.perspectiveapi.com}} a widely used, commercial system for toxicity detection.
Appendix \ref{sec:perspective-details-results} provides a more in-depth description.

We investigate whether \perspectiveAPI scores align with toxicity ratings from workers with specific identities or attitudes, using the 571 posts from our \largescale study.
Specifically, we compare 
the correlations between \perspectiveAPI scores and ratings from annotators, broken down by annotators with different identities (e.g., men and women) or with higher or lower scores on attitude scales (split at the mean).
See Appendix \ref{sec:perspective-details} for details about this methodology.

\begin{figure}[t]
    \centering
    \includegraphics[width=.7\columnwidth]{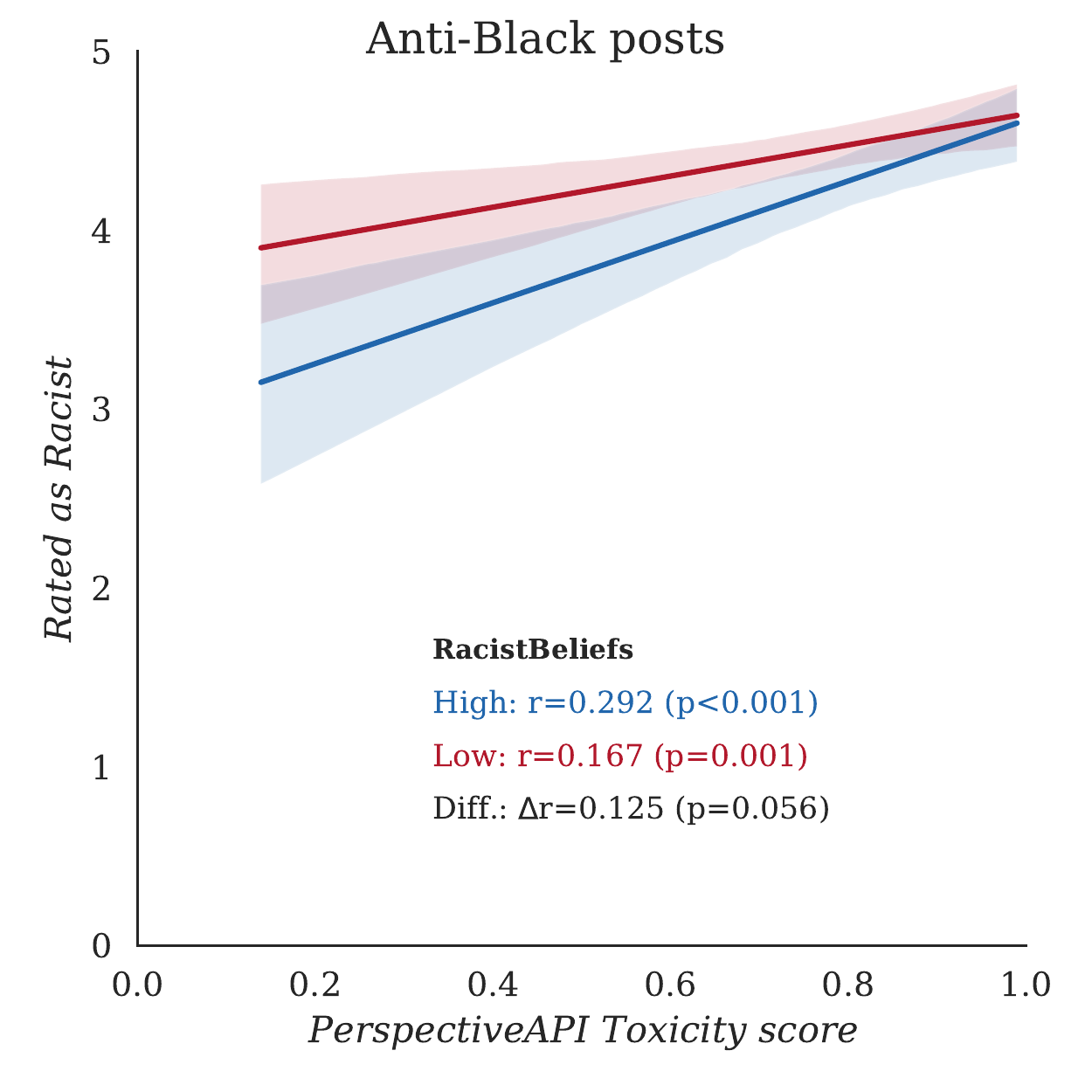}
    \caption{
    Correlation between \perspectiveAPI toxicity scores and racism ratings for \racist posts, broken down by by participants scoring high and low in \racistBeliefs.
    }
    \label{fig:persp-aae-race}
\end{figure}

Our investigation shows that \perspective scores can be significantly more aligned with ratings from certain identities or groups scoring higher or lower on attitude dimensions (see Table~\ref{tab:perspectiveCorrelDiffs} in Appendix \ref{sec:perspective-results}).
Our most salient results show that for \racist posts, \perspective scores are somewhat significantly more aligned with racism ratings by annotators who score high in \racistBeliefs ($r_{\text{high}}$ = 0.29, $r_{\text{low}}$ = 0.17, $\Delta r$ = 0.12, $p$ = 0.056; Figure \ref{fig:persp-aae-race}).
Additionally, for \AAE posts, \perspective scores are slightly more correlated with racism ratings by annotators who were women ($\Delta r$ = 0.22, $p$ < 0.001) or white ($\Delta r$ = 0.08, $p$ = 0.07), and who scored higher in \lingPurism ($\Delta r$ = 0.14, $p$ = 0.003) or \traditionalism ($\Delta r$ = 0.10, $p$ = 0.030).

Overall, our findings indicate that \perspectiveAPI toxicity score predictions align with specific viewpoints or ideologies, depending on the text category. 
Particularly, it seems that the API underestimates the toxicity of \racist posts in a similar way to annotators who scored higher on the \racistBeliefs scale, 
and aligns more with white annotator's perception of \AAE toxicity (vs. Black annotators).
This corroborate prior findings that show that toxicity detection models inherently encode a specific positionality \cite{cambo2021model} and replicate human biases \cite{Davani2021-md}.

\section{Discussion \& Conclusion}
\label{sec:discussion}

Overall, our analyses showed that perceptions of toxicity are indeed affected by annotators' demographic identities and beliefs (\S\ref{sec:whowhatwhy}).
We found---via a \smallscale study and a \largescale study (\S\ref{sec:study-designs})---several associations when isolating specific text characteristics: \racist (\S\ref{sec:racist-results}), \AAE (\S\ref{sec:AAE-results}), and \vulgarity (\S\ref{sec:vulgar-results}). 
Finally, we showed that a popular toxicity detection system yields toxicity scores that are more aligned with raters with certain attitudes and identities than others (\S\ref{sec:perspective-main}).
We discuss implications of our findings below.


\paragraph{Variation in toxicity ratings in hate speech datasets.}
\newText{In our study we deliberately sought rating of \textit{perceptions} of toxicity of posts by racially and politically diverse participants.
However, many existing hate speech datasets instructed annotators to adhere to detailed definitions of toxicity \cite{davidson2017automated,founta2018large}, and some even selected crowdworkers for their liberal ideology \cite[][]{waseem2016you,sap2020socialbiasframes,vidgen-etal-2021-learning}.
While those annotation setups and annotator homogeneity could cause less variation in toxicity annotations of \racist, \AAE, and \vulgar posts, there is still empirical evidence of anti-\AAE racial biases in some of these datasets \cite{sap-etal-2019-risk,davidson-etal-2019-racial}.

Given the large variation in perceptions of toxicity that we showed and the implicit encoding of perspectives by toxicity models, we recommend researchers and dataset creators investigate and report annotator attitudes and demographics;
researchers could collect attitude scores based on relevant social science research, perhaps in lightweight format as done in our \largescale study, and report those scores along with the dataset \cite[e.g., in datasheets;][]{gebru2018datasheets}.
}

\paragraph{Contextualize toxicity predictions in social variables.}
As shown in our results and previous studies \cite[e.g.,][]{waseem2016you,Ross2017measuring,waseem2021disembodied}, determining what is toxic is subjective.
However, given this subjectivity, the open question remains: \textit{whose perspective should be considered when using toxicity detection models?} 
To try answering this question, we urge researchers and practitioners to consider all stakeholders and end users on which toxicity detection systems might be deployed \cite[e.g., through human-centered design methods;][]{sanders2002user,friedman2008value,hovy-yang-2021-importance}.
\newText{While currently, the decision of content moderation often solely lies in the hands of the platforms, we encourage the exploration of alternative solutions \cite[e.g., community fact checkers, digital juries;][]{Maronikolakis2022-xr,Gordon2022-uo}.}

In general, we urge people to embrace that each design decision has socio-political implications \cite{green2020political,cambo2021model}, and encourage them to develop technologies to shift power to the targets of oppression \cite{blodgett-etal-2020-language,Kalluri2020-jk,Birhane2021-ui}.
Finally, given the increasingly essential role of online platforms in people's daily lives \cite{Rahman2017-la}, we echo calls for policy regulating online spaces and toxicity detection algorithms \cite{Jiang2020-qv,Benesch2020-wf,McGuffie2020radicalization,Gillespie2020-wd}.

\paragraph{Beyond toxicity classification: modeling distributions and generating explanations.}
Our findings on the subjectivity of the toxicity detection tasks suggests that standard approaches of obtaining binary (or even $n$-ary) labels of toxicity and averaging them into a majority vote are inadequate.
Instead, researchers could consider modeling the distribution or variation in toxicity labels with respect to individual annotators \cite{geva-etal-2019-modeling,fornaciari-etal-2021-beyond,Davani2021-md} or to specific identities or beliefs.

But, perhaps more importantly, we encourage re-thinking the toxicity detection paradigm altogether.
With the goal to assist human content moderators,\footnote{Note that while content moderation can induce significant psyhcological harms in moderators \cite{Roberts2017-rp,Steiger2021-ka}, full automation also has significant risks.}
creating systems that explain biased implications of posts could be more helpful than opaque toxicity scores
Thus, we advocate for moving away from classification frameworks, and towards more nuanced, holistic, and explainable frameworks for inferring the desired concepts of toxicity and social biases \cite[e.g., Social Bias Frames;][]{sap2020socialbiasframes}.

\paragraph{Limitations and open questions.}
\newText{
Our work had several limitations and raised several open research questions, some of which we outline below.
First, our particular choices of attitudes and scales could affect our results; other scales \cite[e.g., ][for measuring empathy]{gerdes2011measuring} as well as other psychological variables (e.g., propensity to volunteer or to value dignity) could be studied in the context of toxicity perceptions.
Additionally, the automatic \AAE detector in the \largescale study could have induced data selection biases, despite being strongly correlated with race-aware dialect detection (as noted in \cref{footnote:aae-detector-validity}).
Furthermore, our analysis of the attitudes encoded in the \perspectiveAPI in \S\ref{sec:perspective-main} was merely a pilot study; we hope future work will explore more in-depth methods for assess model positionality.

While our study focused on racial discrimination by studying \AAE and \racist posts, future work should explore other axes of discrimination (e.g., sexism, homophobia, ableism, etc.).
Additionally, our study focused only on U.S.-centric perspectives; we hope researchers will explore variations in toxicity perceptions in other cultural contexts (e.g., variations based on caste in India).
}

\bibliographystyle{acl_natbib}
\bibliography{referencesRebibed}

\begin{thebibliography}{110}
\expandafter\ifx\csname natexlab\endcsname\relax\def\natexlab#1{#1}\fi

\bibitem[{Akhtar et~al.(2021)Akhtar, Basile, and Patti}]{akhtar2021whose}
Sohail Akhtar, Valerio Basile, and Viviana Patti. 2021.
\newblock \href {https://arxiv.org/abs/2106.15896} {Whose opinions matter?
  perspective-aware models to identify opinions of hate speech victims in
  abusive language detection}.
\newblock ArXiv preprint arXiv:2106.15896.

\bibitem[{Al~Kuwatly et~al.(2020)Al~Kuwatly, Wich, and
  Groh}]{al-kuwatly-etal-2020-identifying}
Hala Al~Kuwatly, Maximilian Wich, and Georg Groh. 2020.
\newblock \href {https://aclanthology.org/2020.alw-1.21} {Identifying and
  measuring annotator bias based on annotators{'} demographic characteristics}.
\newblock In \emph{Proceedings of the Fourth Workshop on Online Abuse and
  Harms}.

\bibitem[{Are(2020)}]{Are2020-ej}
Carolina Are. 2020.
\newblock \href
  {https://www.tandfonline.com/doi/abs/10.1080/14680777.2020.1783805?journalCode=rfms20}
  {How instagram's algorithm is censoring women and vulnerable users but
  helping online abusers}.
\newblock \emph{Feminist media studies}, 20(5).

\bibitem[{Arhin et~al.(2021)Arhin, Baldini, Wei, Ramamurthy, and
  Singh}]{Arhin2021-wa}
Kofi Arhin, Ioana Baldini, Dennis Wei, Karthikeyan~Natesan Ramamurthy, and
  Moninder Singh. 2021.
\newblock \href {http://arxiv.org/abs/2112.03529} {{Ground-Truth}, whose truth?
  -- examining the challenges with annotating toxic text datasets}.

\bibitem[{Barocas et~al.(2017)Barocas, Crawford, Shapiro, and
  Wallach}]{Barocas2017-bh}
Solon Barocas, Kate Crawford, Aaron Shapiro, and Hanna Wallach. 2017.
\newblock \href
  {http://meetings.sigcis.org/uploads/6/3/6/8/6368912/program.pdf} {The problem
  with bias: Allocative versus representational harms in machine learning}.
\newblock In \emph{{SIGCIS}}.

\bibitem[{Beneke and Cheatham(2015)}]{Beneke2015-to}
Margaret Beneke and Gregory~A Cheatham. 2015.
\newblock Speaking up for african american english: Equity and inclusion in
  early childhood settings.
\newblock \emph{Early Childhood Education Journal}, 43(2).

\bibitem[{Benesch(2020)}]{Benesch2020-wf}
Susan Benesch. 2020.
\newblock \href
  {https://dangerousspeech.org/wp-content/uploads/2020/07/Proposals-for-Improved-Regulation-of-Harmful-Online-Content-Formatted-v5.2.2.pdf}
  {Proposals for improved regulation of harmful online content}.
\newblock Technical report.

\bibitem[{Birhane(2021)}]{Birhane2021-ui}
Abeba Birhane. 2021.
\newblock \href
  {https://www.sciencedirect.com/science/article/pii/S2666389921000155}
  {Algorithmic injustice: a relational ethics approach}.
\newblock \emph{Patterns (New York, N.Y.)}, 2(2).

\bibitem[{Blake and Cutler(2003)}]{Blake2003-ct}
Ren{\'e}e Blake and Cecilia Cutler. 2003.
\newblock {AAE} and variation in teachers' attitudes: A question of school
  philosophy?
\newblock \emph{Linguistics and education}, 14(2).

\bibitem[{Blodgett et~al.(2020)Blodgett, Barocas, Daum{\'e}~III, and
  Wallach}]{blodgett-etal-2020-language}
Su~Lin Blodgett, Solon Barocas, Hal Daum{\'e}~III, and Hanna Wallach. 2020.
\newblock \href {https://aclanthology.org/2020.acl-main.485} {Language
  (technology) is power: A critical survey of {``}bias{''} in {NLP}}.
\newblock In \emph{Proceedings of the 58th Annual Meeting of the Association
  for Computational Linguistics}.

\bibitem[{Blodgett et~al.(2016)Blodgett, Green, and
  O{'}Connor}]{blodgett2016demographic}
Su~Lin Blodgett, Lisa Green, and Brendan O{'}Connor. 2016.
\newblock \href {https://aclanthology.org/D16-1120} {Demographic dialectal
  variation in social media: A case study of {A}frican-{A}merican {E}nglish}.
\newblock In \emph{Proceedings of the 2016 Conference on Empirical Methods in
  Natural Language Processing}.

\bibitem[{Bonilla-Silva(2006)}]{bonilla2006colorblindRacism}
Eduardo Bonilla-Silva. 2006.
\newblock \emph{Racism without racists: Color-blind racism and the persistence
  of racial inequality in the United States}.

\bibitem[{Bouchard~Jr. and McGue(2003)}]{bourchard2003genetic}
Thomas~J. Bouchard~Jr. and Matt McGue. 2003.
\newblock \href
  {http://arxiv.org/abs/https://onlinelibrary.wiley.com/doi/pdf/10.1002/neu.10160}
  {Genetic and environmental influences on human psychological differences}.
\newblock \emph{Journal of Neurobiology}, 54(1).

\bibitem[{Breitfeller et~al.(2019)Breitfeller, Ahn, Jurgens, and
  Tsvetkov}]{breitfeller-etal-2019-finding}
Luke Breitfeller, Emily Ahn, David Jurgens, and Yulia Tsvetkov. 2019.
\newblock \href {https://aclanthology.org/D19-1176} {Finding microaggressions
  in the wild: A case for locating elusive phenomena in social media posts}.
\newblock In \emph{Proceedings of the 2019 Conference on Empirical Methods in
  Natural Language Processing and the 9th International Joint Conference on
  Natural Language Processing (EMNLP-IJCNLP)}.

\bibitem[{Burnham et~al.(2018)Burnham, Le, and Piedmont}]{burnham2018mturk}
Martin~J Burnham, Yen~K Le, and Ralph~L Piedmont. 2018.
\newblock Who is mturk? personal characteristics and sample consistency of
  these online workers.
\newblock \emph{Mental Health, Religion \& Culture}, 21(9-10).

\bibitem[{Cambo(2021)}]{cambo2021model}
Scott~Allen Cambo. 2021.
\newblock \emph{Model Positionality: A Novel Framework for Data Science with
  Subjective Target Concepts}.
\newblock Ph.D. thesis, Northwestern University.

\bibitem[{Carter and Murphy(2015)}]{Carter2015-lc}
Evelyn~R Carter and Mary~C Murphy. 2015.
\newblock Group-based differences in perceptions of racism: What counts, to
  whom, and why?
\newblock \emph{Social and personality psychology compass}, 9(6).

\bibitem[{Champion et~al.(2012)Champion, Cobb-Roberts, and
  Bland-Stewart}]{Champion2012-nv}
Tempii~B Champion, Deirdre Cobb-Roberts, and Linda Bland-Stewart. 2012.
\newblock Future educators' perceptions of african american vernacular english
  ({AAVE}).
\newblock \emph{Online Journal of Education Research}, 1(5).

\bibitem[{Cleland(2014)}]{Cleland2014racism}
Jamie Cleland. 2014.
\newblock Racism, football fans, and online message boards: How social media
  has added a new dimension to racist discourse in {English} football.
\newblock \emph{J. Sport Soc. Issues}, 38(5).

\bibitem[{Cole(1996)}]{cole1996racist}
D~Cole. 1996.
\newblock Racist speech should be protected by the constitution.
\newblock \emph{Hate crimes}.

\bibitem[{Cowan and Khatchadourian(2003)}]{cowan2003empathy}
Gloria Cowan and D{\'e}sir{\'e}e Khatchadourian. 2003.
\newblock \href {https://journals.sagepub.com/doi/abs/10.1111/1471-6402.00110}
  {Empathy, ways of knowing, and interdependence as mediators of gender
  differences in attitudes toward hate speech and freedom of speech}.
\newblock \emph{Psychology of Women Quarterly}, 27(4).

\bibitem[{Cowan et~al.(2002)Cowan, Resendez, Marshall, and
  Quist}]{Cowan2002-at}
Gloria Cowan, Miriam Resendez, Elizabeth Marshall, and Ryan Quist. 2002.
\newblock \href
  {https://www.ojp.gov/ncjrs/virtual-library/abstracts/hate-speech-and-constitutional-protection-priming-values-equality}
  {Hate speech and constitutional protection: Priming values of equality and
  freedom}.
\newblock \emph{The Journal of social issues}, 58(2).

\bibitem[{Croom(2011)}]{Croom2011slurs}
Adam~M Croom. 2011.
\newblock Slurs.
\newblock \emph{Language Sciences}, 33(3).

\bibitem[{Davani et~al.(2021)Davani, Atari, Kennedy, and
  Dehghani}]{Davani2021-md}
Aida~Mostafazadeh Davani, Mohammad Atari, Brendan Kennedy, and Morteza
  Dehghani. 2021.
\newblock \href {https://arxiv.org/abs/2110.14839} {Hate speech classifiers
  learn human-like social stereotypes}.

\bibitem[{Davidson et~al.(2019)Davidson, Bhattacharya, and
  Weber}]{davidson-etal-2019-racial}
Thomas Davidson, Debasmita Bhattacharya, and Ingmar Weber. 2019.
\newblock \href {https://aclanthology.org/W19-3504} {Racial bias in hate speech
  and abusive language detection datasets}.
\newblock In \emph{Proceedings of the Third Workshop on Abusive Language
  Online}.

\bibitem[{Davidson et~al.(2017)Davidson, Warmsley, Macy, and
  Weber}]{davidson2017automated}
Thomas Davidson, Dana Warmsley, Michael Macy, and Ingmar Weber. 2017.
\newblock Automated hate speech detection and the problem of offensive
  language.
\newblock In \emph{ICWSM}.

\bibitem[{DeFrank and Kahlbaugh(2019)}]{defrank2019language}
Melanie DeFrank and Patricia Kahlbaugh. 2019.
\newblock \href
  {https://journals.sagepub.com/doi/full/10.1177/0261927X18758143} {Language
  choice matters: When profanity affects how people are judged}.
\newblock \emph{Journal of Language and Social Psychology}, 38(1).

\bibitem[{Dinan et~al.(2019)Dinan, Humeau, Chintagunta, and
  Weston}]{dinan-etal-2019-build}
Emily Dinan, Samuel Humeau, Bharath Chintagunta, and Jason Weston. 2019.
\newblock \href {https://aclanthology.org/D19-1461} {Build it break it fix it
  for dialogue safety: Robustness from adversarial human attack}.
\newblock In \emph{Proceedings of the 2019 Conference on Empirical Methods in
  Natural Language Processing and the 9th International Joint Conference on
  Natural Language Processing (EMNLP-IJCNLP)}.

\bibitem[{Downs and Cowan(2012)}]{Downs2012-lu}
Daniel~M Downs and Gloria Cowan. 2012.
\newblock \href
  {https://onlinelibrary.wiley.com/doi/abs/10.1111/j.1559-1816.2012.00902.x}
  {Predicting the importance of freedom of speech and the perceived harm of
  hate speech}.
\newblock \emph{Journal of applied social psychology}, 42(6).

\bibitem[{Duckitt and Fisher(2003)}]{duckitt2003impact}
John Duckitt and Kirstin Fisher. 2003.
\newblock \href
  {https://onlinelibrary.wiley.com/doi/abs/10.1111/0162-895X.00322} {The impact
  of social threat on worldview and ideological attitudes}.
\newblock \emph{Political Psychology}, 24(1).

\bibitem[{Dynel(2012)}]{Dynel2012Swearing}
Marta Dynel. 2012.
\newblock \href
  {https://publicaciones.unirioja.es/ojs/index.php/jes/article/view/179}
  {Swearing methodologically : the (im)politeness of expletives in anonymous
  commentaries on youtube}.
\newblock \emph{Journal of English Studies}, 10(0).

\bibitem[{Edwards(2004)}]{edwards2004african}
Walter~F Edwards. 2004.
\newblock African american vernacular english: phonology.
\newblock In \emph{A Handbook of Varieties of English: Morphology and syntax}.

\bibitem[{Elers and Jayan(2020)}]{Elers2020-ne}
Christine~Helen Elers and Pooja Jayan. 2020.
\newblock ``this is us'': Free speech embedded in whiteness, racism and
  coloniality in aotearoa, new zealand.
\newblock \emph{First Amendment Studies}, 54(2).

\bibitem[{Fiske(1993)}]{Fiske1993controlling}
S~T Fiske. 1993.
\newblock Controlling other people. the impact of power on stereotyping.
\newblock \emph{The American psychologist}, 48(6).

\bibitem[{Fornaciari et~al.(2021)Fornaciari, Uma, Paun, Plank, Hovy, and
  Poesio}]{fornaciari-etal-2021-beyond}
Tommaso Fornaciari, Alexandra Uma, Silviu Paun, Barbara Plank, Dirk Hovy, and
  Massimo Poesio. 2021.
\newblock \href {https://aclanthology.org/2021.naacl-main.204} {Beyond black
  {\&} white: Leveraging annotator disagreement via soft-label multi-task
  learning}.
\newblock In \emph{Proceedings of the 2021 Conference of the North American
  Chapter of the Association for Computational Linguistics: Human Language
  Technologies}.

\bibitem[{Founta et~al.(2018)Founta, Djouvas, Chatzakou, Leontiadis, Blackburn,
  Stringhini, Vakali, Sirivianos, and Kourtellis}]{founta2018large}
Antigoni~Maria Founta, Constantinos Djouvas, Despoina Chatzakou, Ilias
  Leontiadis, Jeremy Blackburn, Gianluca Stringhini, Athena Vakali, Michael
  Sirivianos, and Nicolas Kourtellis. 2018.
\newblock \href {https://arxiv.org/abs/1802.00393} {Large scale crowdsourcing
  and characterization of twitter abusive behavior}.
\newblock In \emph{ICWSM}.

\bibitem[{Friedman et~al.(2008)Friedman, Kahn, and Borning}]{friedman2008value}
Batya Friedman, Peter~H Kahn, and Alan Borning. 2008.
\newblock \href
  {https://vsdesign.org/publications/pdf/non-scan-vsd-and-information-systems.pdf}
  {Value sensitive design and information systems}.
\newblock \emph{The handbook of information and computer ethics}.

\bibitem[{Gaither et~al.(2015)Gaither, Cohen-Goldberg, Gidney, and
  Maddox}]{gaither2015sounding}
Sarah~E Gaither, Ariel~M Cohen-Goldberg, Calvin~L Gidney, and Keith~B Maddox.
  2015.
\newblock \href
  {https://internal-journal.frontiersin.org/articles/10.3389/fpsyg.2015.00457/full}
  {Sounding black or white: Priming identity and biracial speech}.
\newblock \emph{Frontiers in Psychology}, 6.

\bibitem[{Galinsky et~al.(2013)Galinsky, Wang, Whitson, Anicich, Hugenberg, and
  Bodenhausen}]{Galinsky2013-rw}
Adam~D Galinsky, Cynthia~S Wang, Jennifer~A Whitson, Eric~M Anicich, Kurt
  Hugenberg, and Galen~V Bodenhausen. 2013.
\newblock The reappropriation of stigmatizing labels: the reciprocal
  relationship between power and self-labeling.
\newblock \emph{Psychol. Sci.}, 24(10).

\bibitem[{Gavrilets and Fortunato(2014)}]{gavrilets2014solution}
Sergey Gavrilets and Laura Fortunato. 2014.
\newblock A solution to the collective action problem in between-group conflict
  with within-group inequality.
\newblock \emph{Nature communications}, 5(1).

\bibitem[{Gebru et~al.(2018)Gebru, Morgenstern, Vecchione, Vaughan, Wallach,
  III, and Crawford}]{gebru2018datasheets}
Timnit Gebru, Jamie Morgenstern, Briana Vecchione, Jennifer~Wortman Vaughan,
  Hanna Wallach, Hal~Daumé III, and Kate Crawford. 2018.
\newblock \href {https://arxiv.org/abs/1803.09010} {Datasheets for datasets}.
\newblock In \emph{FAccT*}.

\bibitem[{Gerdes et~al.(2011)Gerdes, Lietz, and Segal}]{gerdes2011measuring}
Karen~E Gerdes, Cynthia~A Lietz, and Elizabeth~A Segal. 2011.
\newblock Measuring empathy in the 21st century: Development of an empathy
  index rooted in social cognitive neuroscience and social justice.
\newblock \emph{Social Work Research}, 35(2).

\bibitem[{Geva et~al.(2019)Geva, Goldberg, and
  Berant}]{geva-etal-2019-modeling}
Mor Geva, Yoav Goldberg, and Jonathan Berant. 2019.
\newblock \href {https://aclanthology.org/D19-1107} {Are we modeling the task
  or the annotator? an investigation of annotator bias in natural language
  understanding datasets}.
\newblock In \emph{Proceedings of the 2019 Conference on Empirical Methods in
  Natural Language Processing and the 9th International Joint Conference on
  Natural Language Processing (EMNLP-IJCNLP)}.

\bibitem[{Gillborn(2009)}]{Gillborn2009-rv}
David Gillborn. 2009.
\newblock Risk-free racism: Whiteness and so-called free speech.
\newblock \emph{Wake Forest law review}, 44.

\bibitem[{Gillespie et~al.(2020)Gillespie, Aufderheide, Carmi, Gerrard, Gorwa,
  Matamoros-Fernandez, Roberts, Sinnreich, and West}]{Gillespie2020-wd}
Tarleton Gillespie, Patricia Aufderheide, Elinor Carmi, Ysabel Gerrard, Robert
  Gorwa, Ariadna Matamoros-Fernandez, Sarah~T Roberts, Aram Sinnreich, and
  Sarah~Myers West. 2020.
\newblock \href
  {https://policyreview.info/articles/analysis/expanding-debate-about-content-moderation-scholarly-research-agendas-coming-policy}
  {Expanding the debate about content moderation: Scholarly research agendas
  for the coming policy debates}.
\newblock \emph{Internet Policy Review}, 9(4).

\bibitem[{Goldstein and Hall(2017)}]{Goldstein2017-mx}
Donna~M Goldstein and Kira Hall. 2017.
\newblock Postelection surrealism and nostalgic racism in the hands of donald
  trump.
\newblock \emph{HAU: Journal of Ethnographic Theory}, 7(1).

\bibitem[{Gordon et~al.(2022)Gordon, Lam, Park, Patel, Hancock, Hashimoto, and
  Bernstein}]{Gordon2022-uo}
Mitchell~L Gordon, Michelle~S Lam, Joon~Sung Park, Kayur Patel, Jeffrey~T
  Hancock, Tatsunori Hashimoto, and Michael~S Bernstein. 2022.
\newblock \href {http://arxiv.org/abs/2202.02950} {Jury learning: Integrating
  dissenting voices into machine learning models}.
\newblock In \emph{{CHI}}.

\bibitem[{Green(2020)}]{green2020political}
Ben Green. 2020.
\newblock \href {http://dx.doi.org/10.2139/ssrn.3658431} {Data science as
  political action: Grounding data science in a politics of justice}.
\newblock \emph{SSRN Electronic Journal}.

\bibitem[{Green(2002)}]{Green2002aae}
Lisa Green. 2002.
\newblock \emph{African American English: A Linguistic Introduction}, 8.3.2002
  edition edition.

\bibitem[{Greengross and Miller(2008)}]{Greengross2008selfdeprecating}
Gil Greengross and Geoffrey~F Miller. 2008.
\newblock Dissing oneself versus dissing rivals: effects of status,
  personality, and sex on the {Short-Term} and {Long-Term} attractiveness of
  {Self-Deprecating} and {Other-Deprecating} humor.
\newblock \emph{Evolutionary Psychology}, 6(3).

\bibitem[{Han and Tsvetkov(2020)}]{han-tsvetkov-2020-fortifying}
Xiaochuang Han and Yulia Tsvetkov. 2020.
\newblock \href {https://aclanthology.org/2020.emnlp-main.622} {Fortifying
  toxic speech detectors against veiled toxicity}.
\newblock In \emph{Proceedings of the 2020 Conference on Empirical Methods in
  Natural Language Processing (EMNLP)}.

\bibitem[{Hilliard(1997)}]{Hilliard1997-vx}
A~G Hilliard. 1997.
\newblock Language, culture and the assessment of african american children.
\newblock \emph{Assessment for equity and inclusion: Embracing all our
  children}.

\bibitem[{Hovy and Yang(2021)}]{hovy-yang-2021-importance}
Dirk Hovy and Diyi Yang. 2021.
\newblock \href {https://aclanthology.org/2021.naacl-main.49} {The importance
  of modeling social factors of language: Theory and practice}.
\newblock In \emph{Proceedings of the 2021 Conference of the North American
  Chapter of the Association for Computational Linguistics: Human Language
  Technologies}.

\bibitem[{Huff and Tingley(2015)}]{huff2015these}
Connor Huff and Dustin Tingley. 2015.
\newblock \href {https://journals.sagepub.com/doi/10.1177/2053168015604648}
  {“who are these people?” evaluating the demographic characteristics and
  political preferences of mturk survey respondents}.
\newblock \emph{Research \& Politics}, 2(3).

\bibitem[{Ilbury(2020)}]{Ilbury2020-mb}
Christian Ilbury. 2020.
\newblock \href {https://onlinelibrary.wiley.com/doi/10.1111/josl.12366}
  {``sassy queens'': Stylistic orthographic variation in twitter and the
  enregisterment of {AAVE}}.
\newblock \emph{Journal of sociolinguistics}, 24(2).

\bibitem[{Jay(2018)}]{Jay2018-yt}
Timothy Jay. 2018.
\newblock Swearing, moral order, and online communication.
\newblock \emph{Journal of Language Aggression and Conflict}, 6(1).

\bibitem[{Jay and Janschewitz(2008)}]{jay2008pragmatics}
Timothy Jay and Kristin Janschewitz. 2008.
\newblock \href
  {https://www.degruyter.com/document/doi/10.1515/JPLR.2008.013/html} {The
  pragmatics of swearing}.
\newblock \emph{Journal of Politeness Research}, 4.

\bibitem[{Jernudd and Shapiro(1989)}]{jernudd1989politics}
Bj{\"o}rn~H Jernudd and Michael~J Shapiro. 1989.
\newblock \emph{The politics of language purism}.

\bibitem[{Jiang(2020)}]{Jiang2020-qv}
J~A Jiang. 2020.
\newblock Identifying and addressing design and policy challenges in online
  content moderation.
\newblock \emph{Extended Abstracts of the 2020 CHI Conference on}.

\bibitem[{Jiang et~al.(2021)Jiang, Scheuerman, Fiesler, and
  Brubaker}]{Jiang2021understanding}
Jialun~Aaron Jiang, Morgan~Klaus Scheuerman, Casey Fiesler, and Jed~R Brubaker.
  2021.
\newblock Understanding international perceptions of the severity of harmful
  content online.
\newblock \emph{PloS one}, 16(8).

\bibitem[{Johnson et~al.(2022)Johnson, Mattan, Flores, Lauharatanahirun, and
  Falk}]{Johnson2022-mp}
Darin~G Johnson, Bradley~D Mattan, Nelson Flores, Nina Lauharatanahirun, and
  Emily~B Falk. 2022.
\newblock \href {https://doi.org/10.1007/s42761-021-00072-8}
  {{Social-Cognitive} and affective antecedents of code switching and the
  consequences of linguistic racism for black people and people of color}.
\newblock \emph{Affective Science}, 3(1).

\bibitem[{Johnson and Tamney(2001)}]{johnson2001social}
Stephen~D Johnson and Joseph~B Tamney. 2001.
\newblock Social traditionalism and economic conservatism: Two conservative
  political ideologies in the united states.
\newblock \emph{The Journal of social psychology}, 141(2).

\bibitem[{Kalluri(2020)}]{Kalluri2020-jk}
Pratyusha Kalluri. 2020.
\newblock \href {https://www.nature.com/articles/d41586-020-02003-2} {Don't ask
  if artificial intelligence is good or fair, ask how it shifts power}.
\newblock \emph{Nature}, 583(7815).

\bibitem[{Knuckey(2005)}]{knuckey2005new}
Jonathan Knuckey. 2005.
\newblock A new front in the culture war? moral traditionalism and voting
  behavior in us house elections.
\newblock \emph{American Politics Research}, 33(5).

\bibitem[{Loepp and Kelly(2020)}]{loepp2020distinction}
Eric Loepp and Jarrod~T Kelly. 2020.
\newblock Distinction without a difference? an assessment of mturk worker
  types.
\newblock \emph{Research \& Politics}, 7(1).

\bibitem[{Lucks(2020)}]{Lucks2020-ri}
Daniel~S Lucks. 2020.
\newblock \emph{Reconsidering Reagan: Racism, Republicans, and the Road to
  Trump}.

\bibitem[{Maronikolakis et~al.(2022)Maronikolakis, Wisiorek, Nann, Jabbar,
  Udupa, and Sch{\"u}tze}]{Maronikolakis2022-xr}
Antonis Maronikolakis, Axel Wisiorek, Leah Nann, Haris Jabbar, Sahana Udupa,
  and Hinrich Sch{\"u}tze. 2022.
\newblock \href
  {https://raw.githubusercontent.com/antmarakis/antmarakis.github.io/master/files/xtremespeech.pdf}
  {Listening to affected communities to define extreme speech: Dataset and
  experiments}.
\newblock In \emph{{ACL} 2022 Findings}.

\bibitem[{McConahay(1986)}]{mcconahay1986modern}
John~B McConahay. 1986.
\newblock \href {https://psycnet.apa.org/record/1986-98698-004} {Modern racism,
  ambivalence, and the modern racism scale}.
\newblock In \emph{Prejudice, discrimination, and racism}, volume 337.

\bibitem[{McGuffie and Newhouse(2020)}]{McGuffie2020radicalization}
Kris McGuffie and Alex Newhouse. 2020.
\newblock \href {http://arxiv.org/abs/2009.06807} {The radicalization risks of
  gpt-3 and advanced neural language models}.

\bibitem[{Muddiman(2021)}]{Muddiman2021-zz}
Ashley Muddiman. 2021.
\newblock Conservatives and incivility.
\newblock In \emph{Conservative Political Communication}.

\bibitem[{Nadal(2018)}]{Nadal2018-iz}
Kevin~L Nadal. 2018.
\newblock \emph{Microaggressions and traumatic stress: Theory, research, and
  clinical treatment}.

\bibitem[{Nelson et~al.(2013)Nelson, Adams, and Salter}]{nelson2013marley}
Jessica~C Nelson, Glenn Adams, and Phia~S Salter. 2013.
\newblock The marley hypothesis: Denial of racism reflects ignorance of
  history.
\newblock \emph{Psychological science}, 24(2).

\bibitem[{Norton and Sommers(2011)}]{Norton2011-gw}
Michael~I Norton and Samuel~R Sommers. 2011.
\newblock \href {https://www.hbs.edu/faculty/Pages/item.aspx?num=40027} {Whites
  see racism as a {Zero-Sum} game that they are now losing}.
\newblock \emph{Perspectives on psychological science: a journal of the
  Association for Psychological Science}, 6(3).

\bibitem[{O'Keeffe et~al.(2011)O'Keeffe, Clarke-Pearson, and {Council on
  Communications and Media}}]{OKeeffe2011impact}
Gwenn~Schurgin O'Keeffe, Kathleen Clarke-Pearson, and {Council on
  Communications and Media}. 2011.
\newblock The impact of social media on children, adolescents, and families.
\newblock \emph{Pediatrics}, 127(4).

\bibitem[{Pavlick and Kwiatkowski(2019)}]{Pavlick2019-tc}
Ellie Pavlick and Tom Kwiatkowski. 2019.
\newblock \href {https://aclanthology.org/Q19-1043} {Inherent disagreements in
  human textual inferences}.
\newblock \emph{Transactions of the Association for Computational Linguistics},
  7.

\bibitem[{Poteat and Spanierman(2012)}]{poteat2012modern}
V~Paul Poteat and Lisa~B Spanierman. 2012.
\newblock \href
  {https://www.tandfonline.com/doi/abs/10.1080/00224545.2012.700966} {Modern
  racism attitudes among white students: The role of dominance and
  authoritarianism and the mediating effects of racial color-blindness}.
\newblock \emph{The Journal of Social Psychology}, 152(6).

\bibitem[{Prabhakaran et~al.(2021)Prabhakaran, Davani, and
  D{\'\i}az}]{Prabhakaran2021-gx}
Vinodkumar Prabhakaran, Aida~Mostafazadeh Davani, and Mark D{\'\i}az. 2021.
\newblock \href {https://arxiv.org/abs/2110.05699} {On releasing
  {Annotator-Level} labels and information in datasets}.
\newblock In \emph{Proc. of LAW-DMR workshop at EMNLP}.

\bibitem[{Pulos et~al.(2004)Pulos, Elison, and Lennon}]{pulos2004hierarchical}
Steven Pulos, Jeff Elison, and Randy Lennon. 2004.
\newblock \href {https://www.sbp-journal.com/index.php/sbp/article/view/1336}
  {The hierarchical structure of the interpersonal reactivity index}.
\newblock \emph{Social behavior and personality}, 32(4).

\bibitem[{Rahman(2008)}]{Rahman2008-lu}
Jacquelyn Rahman. 2008.
\newblock \href
  {https://read.dukeupress.edu/american-speech/article/83/2/141/5819/MIDDLE-CLASS-AFRICAN-AMERICANS-REACTIONS-AND}
  {Middle-class african americans: Reactions and attitudes toward african
  american english}.
\newblock \emph{American speech}, 83(2).

\bibitem[{Rahman(2017)}]{Rahman2017-la}
K~Sabeel Rahman. 2017.
\newblock The new utilities: Private power, social infrastructure, and the
  revival of the public utility concept.
\newblock \emph{Cardozo law review}, 39.

\bibitem[{Riar et~al.(2020)Riar, Morschheuser, Hamari, and
  Zarnekow}]{riar2020game}
Marc Riar, Benedikt Morschheuser, Juho Hamari, and R{\"u}diger Zarnekow. 2020.
\newblock How game features give rise to altruism and collective action?
  implications for cultivating cooperation by gamification.
\newblock In \emph{Proceedings of the 53rd Hawaii International Conference on
  System Sciences}.

\bibitem[{Roberts(2017)}]{Roberts2017-rp}
Sarah~T Roberts. 2017.
\newblock \href
  {https://www.theatlantic.com/technology/archive/2017/03/commercial-content-moderation/518796/}
  {Social media's silent filter}.
\newblock \emph{The Atlantic}.

\bibitem[{Rosa(2019)}]{Rosa2019soundingLikeRace}
Jonathan Rosa. 2019.
\newblock \emph{Looking Like a Language, Sounding Like a Race}.

\bibitem[{Rosa and Flores(2017)}]{Rosa2017-ec}
Jonathan Rosa and Nelson Flores. 2017.
\newblock Unsettling race and language: Toward a raciolinguistic perspective.
\newblock \emph{Language In Society}, 46(5).

\bibitem[{Ross et~al.(2017)Ross, Rist, Carbonell, Cabrera, Kurowsky, and
  Wojatzki}]{Ross2017measuring}
Bj{\"o}rn Ross, Michael Rist, Guillermo Carbonell, Benjamin Cabrera, Nils
  Kurowsky, and Michael Wojatzki. 2017.
\newblock \href {https://arxiv.org/abs/1701.08118} {Measuring the reliability
  of hate speech annotations: the case of the european refugee crisis}.
\newblock In \emph{{NLP} 4 {CMC} Workshop}.

\bibitem[{Roussos and Dovidio(2018)}]{Roussos2018-sf}
Gina Roussos and John~F Dovidio. 2018.
\newblock Hate speech is in the eye of the beholder: The influence of racial
  attitudes and freedom of speech beliefs on perceptions of racially motivated
  threats of violence.
\newblock \emph{Social psychological and personality science}, 9(2).

\bibitem[{Sanders(2002)}]{sanders2002user}
Elizabeth Sanders. 2002.
\newblock \emph{From user-centered to participatory design approaches}.

\bibitem[{Sap et~al.(2019)Sap, Card, Gabriel, Choi, and
  Smith}]{sap-etal-2019-risk}
Maarten Sap, Dallas Card, Saadia Gabriel, Yejin Choi, and Noah~A. Smith. 2019.
\newblock \href {https://aclanthology.org/P19-1163} {The risk of racial bias in
  hate speech detection}.
\newblock In \emph{Proceedings of the 57th Annual Meeting of the Association
  for Computational Linguistics}.

\bibitem[{Sap et~al.(2020)Sap, Gabriel, Qin, Jurafsky, Smith, and
  Choi}]{sap2020socialbiasframes}
Maarten Sap, Saadia Gabriel, Lianhui Qin, Dan Jurafsky, Noah~A. Smith, and
  Yejin Choi. 2020.
\newblock \href {https://aclanthology.org/2020.acl-main.486} {Social bias
  frames: Reasoning about social and power implications of language}.
\newblock In \emph{Proceedings of the 58th Annual Meeting of the Association
  for Computational Linguistics}.

\bibitem[{Sapolsky et~al.(2010)Sapolsky, Shafer, and Kaye}]{sapolsky2010rating}
Barry~S. Sapolsky, Daniel~M. Shafer, and Barbara~K. Kaye. 2010.
\newblock \href
  {http://arxiv.org/abs/https://doi.org/10.1080/15205430903359693} {Rating
  offensive words in three television program contexts}.
\newblock \emph{Mass Communication and Society}, 14(1).

\bibitem[{Schaffner(2020)}]{Schaffner2020-jv}
Brian~F Schaffner. 2020.
\newblock The heightened importance of racism and sexism in the 2018 {US}
  midterm elections.
\newblock \emph{British journal of political science}.

\bibitem[{Silva et~al.(2016)Silva, Mondal, Correa, Benevenuto, and
  Weber}]{silva2016analyzing}
Leandro Silva, Mainack Mondal, Denzil Correa, Fabr{\'\i}cio Benevenuto, and
  Ingmar Weber. 2016.
\newblock Analyzing the targets of hate in online social media.
\newblock In \emph{Tenth international AAAI conference on web and social
  media}.

\bibitem[{Silver and Dunlap(1987)}]{silver1987averaging}
N~Clayton Silver and William~P Dunlap. 1987.
\newblock Averaging correlation coefficients: should fisher's z transformation
  be used?
\newblock \emph{Journal of applied psychology}, 72(1).

\bibitem[{Soral et~al.(2018)Soral, Bilewicz, and Winiewski}]{Soral2018-wh}
Wiktor Soral, Micha{\l} Bilewicz, and Miko{\l}aj Winiewski. 2018.
\newblock Exposure to hate speech increases prejudice through desensitization.
\newblock \emph{Aggressive behavior}, 44(2).

\bibitem[{Spears et~al.(1998)}]{spears1998african}
Arthur~K Spears et~al. 1998.
\newblock \href
  {https://books.google.com/books?hl=en&lr=&id=fyFBq7rQgpkC&oi=fnd&pg=PA226&ots=rod7BZdLDQ&sig=yz9TTODBb3elV4f0Uu2v1m8Z7VQ#v=onepage&q&f=false}
  {African-american language use: Ideology and so-called obscenity}.
\newblock \emph{African-American English: Structure, history, and use}.

\bibitem[{Steg et~al.(2014)Steg, Perlaviciute, Van~der Werff, and
  Lurvink}]{steg2014significance}
Linda Steg, Goda Perlaviciute, Ellen Van~der Werff, and Judith Lurvink. 2014.
\newblock The significance of hedonic values for environmentally relevant
  attitudes, preferences, and actions.
\newblock \emph{Environment and behavior}, 46(2).

\bibitem[{Steiger et~al.(2021)Steiger, Bharucha, Venkatagiri, Riedl, and
  Lease}]{Steiger2021-ka}
Miriah Steiger, Timir~J Bharucha, Sukrit Venkatagiri, Martin~J Riedl, and
  Matthew Lease. 2021.
\newblock \href {https://doi.org/10.1145/3411764.3445092} {The psychological
  {Well-Being} of content moderators: The emotional labor of commercial
  moderation and avenues for improving support}.
\newblock In \emph{CHI}, number Article 341 in CHI '21.

\bibitem[{Sterling et~al.(2020)Sterling, Jost, and Bonneau}]{Sterling2020-cq}
Joanna Sterling, John~T Jost, and Richard Bonneau. 2020.
\newblock Political psycholinguistics: A comprehensive analysis of the language
  habits of liberal and conservative social media users.
\newblock \emph{Journal of personality and social psychology}, 118(4).

\bibitem[{Ueland(2020)}]{Ueland2020-pt}
Ane Ueland. 2020.
\newblock \href {https://uis.brage.unit.no/uis-xmlui/handle/11250/2670762}
  {\emph{Language and identity: a study of African American Vernacular English
  and its status in American society}}.
\newblock Ph.D. thesis, University of Stavanger, Norway.

\bibitem[{Vidgen et~al.(2021)Vidgen, Thrush, Waseem, and
  Kiela}]{vidgen-etal-2021-learning}
Bertie Vidgen, Tristan Thrush, Zeerak Waseem, and Douwe Kiela. 2021.
\newblock \href {https://aclanthology.org/2021.acl-long.132} {Learning from the
  worst: Dynamically generated datasets to improve online hate detection}.
\newblock In \emph{Proceedings of the 59th Annual Meeting of the Association
  for Computational Linguistics and the 11th International Joint Conference on
  Natural Language Processing (Volume 1: Long Papers)}.

\bibitem[{Wagstaff(1998)}]{wagstaff1998equity}
Graham~F Wagstaff. 1998.
\newblock Equity, justice, and altruism.
\newblock \emph{Current Psychology}, 17(2).

\bibitem[{Waseem(2016)}]{waseem2016you}
Zeerak Waseem. 2016.
\newblock \href {https://aclanthology.org/W16-5618} {Are you a racist or am {I}
  seeing things? annotator influence on hate speech detection on {T}witter}.
\newblock In \emph{Proceedings of the First Workshop on {NLP} and Computational
  Social Science}.

\bibitem[{Waseem and Hovy(2016)}]{waseem-hovy-2016-hateful}
Zeerak Waseem and Dirk Hovy. 2016.
\newblock \href {https://aclanthology.org/N16-2013} {Hateful symbols or hateful
  people? predictive features for hate speech detection on {T}witter}.
\newblock In \emph{Proceedings of the {NAACL} Student Research Workshop}.

\bibitem[{Waseem et~al.(2021)Waseem, Lulz, Bingel, and
  Augenstein}]{waseem2021disembodied}
Zeerak Waseem, Smarika Lulz, Joachim Bingel, and Isabelle Augenstein. 2021.
\newblock \href {https://arxiv.org/abs/2101.11974} {Disembodied machine
  learning: On the illusion of objectivity in nlp}.
\newblock Anonymous preprint under review.

\bibitem[{White and Crandall(2017)}]{White2017-ad}
Mark~H White and Christian~S Crandall. 2017.
\newblock Freedom of racist speech: Ego and expressive threats.
\newblock \emph{Journal of personality and social psychology}, 113(3).

\bibitem[{Wulczyn et~al.(2017)Wulczyn, Thain, and Dixon}]{wulczyn2017ex}
Ellery Wulczyn, Nithum Thain, and Lucas Dixon. 2017.
\newblock \href {https://doi.org/10.1145/3038912.3052591} {Ex machina: Personal
  attacks seen at scale}.
\newblock In \emph{Proceedings of the 26th International Conference on World
  Wide Web, {WWW} 2017, Perth, Australia, April 3-7, 2017}.

\bibitem[{Yasin(2018)}]{Yasin2018black}
Danyaal Yasin. 2018.
\newblock \href
  {https://www.indexoncensorship.org/2018/09/black-and-banned-who-is-free-speech-for/}
  {Black and banned: Who is free speech for?}
\newblock \url{}.
\newblock Accessed: 2018-12-6.

\bibitem[{Zampieri et~al.(2019)Zampieri, Malmasi, Nakov, Rosenthal, Farra, and
  Kumar}]{Zampieri2019OLID}
Marcos Zampieri, Shervin Malmasi, Preslav Nakov, Sara Rosenthal, Noura Farra,
  and Ritesh Kumar. 2019.
\newblock \href {https://aclanthology.org/N19-1144} {Predicting the type and
  target of offensive posts in social media}.
\newblock In \emph{Proceedings of the 2019 Conference of the North {A}merican
  Chapter of the Association for Computational Linguistics: Human Language
  Technologies, Volume 1 (Long and Short Papers)}.

\bibitem[{Zevallos(2017)}]{Zevallos2017sociologyOfRace}
Zuleyka Zevallos. 2017.
\newblock \href {https://othersociologist.com/sociology-of-race/} {Sociology of
  race}.
\newblock \url{}.
\newblock Accessed: 2021-10-10.

\bibitem[{Zhou et~al.(2021)Zhou, Sap, Swayamdipta, Choi, and
  Smith}]{zhou2021challenges}
Xuhui Zhou, Maarten Sap, Swabha Swayamdipta, Yejin Choi, and Noah Smith. 2021.
\newblock \href {https://aclanthology.org/2021.eacl-main.274} {Challenges in
  automated debiasing for toxic language detection}.
\newblock In \emph{Proceedings of the 16th Conference of the European Chapter
  of the Association for Computational Linguistics: Main Volume}.

\end{thebibliography}

\appendix
\clearpage

\section{Attitude Scales}
\label{app:attitude_scales}

Below, we list the individual items from each of our seven attitude dimensions.
In our \smallscale study, we asked participants all of the items in each scale, but in the larger-scale study, we only asked the bolded items (one per scale).
(R) denotes a reverse coded item.

\subsection{Valuing the freedom of offensive speech (\freeSpeech).}
\label{app:free_speech}
Participants were instructed to rate their level of agreement with the item statements, using a 5-point Likert scale ranging from ``\textit{not at all}'' (1) to ``\textit{very much so}'' (5).
\begin{compactitem}  
    \item People should face consequences for saying something offensive online. (R)
    \item Offensive posts and comments online should be removed from the platforms.
    \item  \textbf{There is no such a thing as hate speech, only ``speech''.} 
    \item  People should be allowed to say things that others may consider offensive.
    \item  It's more important to protect free speech rights than to protect people's feelings when it comes to offensive language.
\end{compactitem}
This scale was taken from \citet{Cowan2002-at}.

\subsection{Perceiving the \harmOfHateSpeech.} 
\label{app:harm_hate_speech}
Participants were instructed to rate their level of agreement with the item statements, using a 5-point Likert scale ranging from ``\textit{not at all}'' (1) to ``\textit{very much so}'' (5).
\begin{compactitem}  
    \item When someone posts something offensive towards minorities online it intimidates and casts fear in the heart of individuals who are part of the minority group.
    \item \textbf{Offensive language encourages discrimination against minorities.}
    \item Offensive language is harmful to minorities. 
\end{compactitem}
This scale was taken from \citet{Cowan2002-at}.

\subsection{Endorsement of \racistBeliefs.} 
\label{app:racism}
Participants were instructed to rate their level of agreement with the item statements, using a 5-point Likert scale ranging from ``\textit{not at all}'' (1) to ``\textit{very much so}'' (5).
\begin{compactitem}  
    \item \textbf{Discrimination against racial minorities is no longer a problem in the United States.}
    \item It is easy to understand the anger of racial minorities people in America. (R)
    \item Racial minorities are getting too demanding in their push for equal rights.
    \item Over the past few years, racial minorities have gotten more economically than they deserve.
    \item Over the past few years, the government and news media have shown more respect to racial minorities than they deserve.
\end{compactitem}
These items form the validated Modern Racism Scale, created by \citet{mcconahay1986modern}.

\subsection{\traditionalism.}
\label{app:traditionalism}
Participants were asked: ``Please tell us how important each of these is as a guiding principle in your life.''
They answered each item on a 5-point Likert scale, ranging from ``\textit{not at all important to me}'' (1) to ``\textit{extremely very important to me}'' (5).
\begin{compactitem}  
    \item Being obedient, dutiful, meeting obligations.
    \item Self-discipline, self-restraint, resistance to temptations.
    \item \textbf{Honoring parents and elders, showing respect.}
    \item Traditions and customs.
\end{compactitem}
This is an abridged version of the traditionalism scale by \citet{bourchard2003genetic}.

\subsection{Language Purism (\lingPurism).} 
\label{app:language_purism}
Participants were instructed to rate their level of agreement with the item statements, using a 5-point Likert scale ranging from ``\textit{not at all}'' (1) to ``\textit{very much so}'' (5).
\begin{compactitem}  
    \item I dislike when people make simple grammar or spelling errors.
    \item It is important to master the English language properly, and not make basic spelling mistakes or misuse a common word.
    \item I am not afraid to correct people when they make simple grammar or spelling errors.
    \item \textbf{There exists such a thing as good proper English.}
\end{compactitem}
This scale was created by the authors.

\begin{figure*}[t]
    \centering
    \includegraphics[width=.9\textwidth]{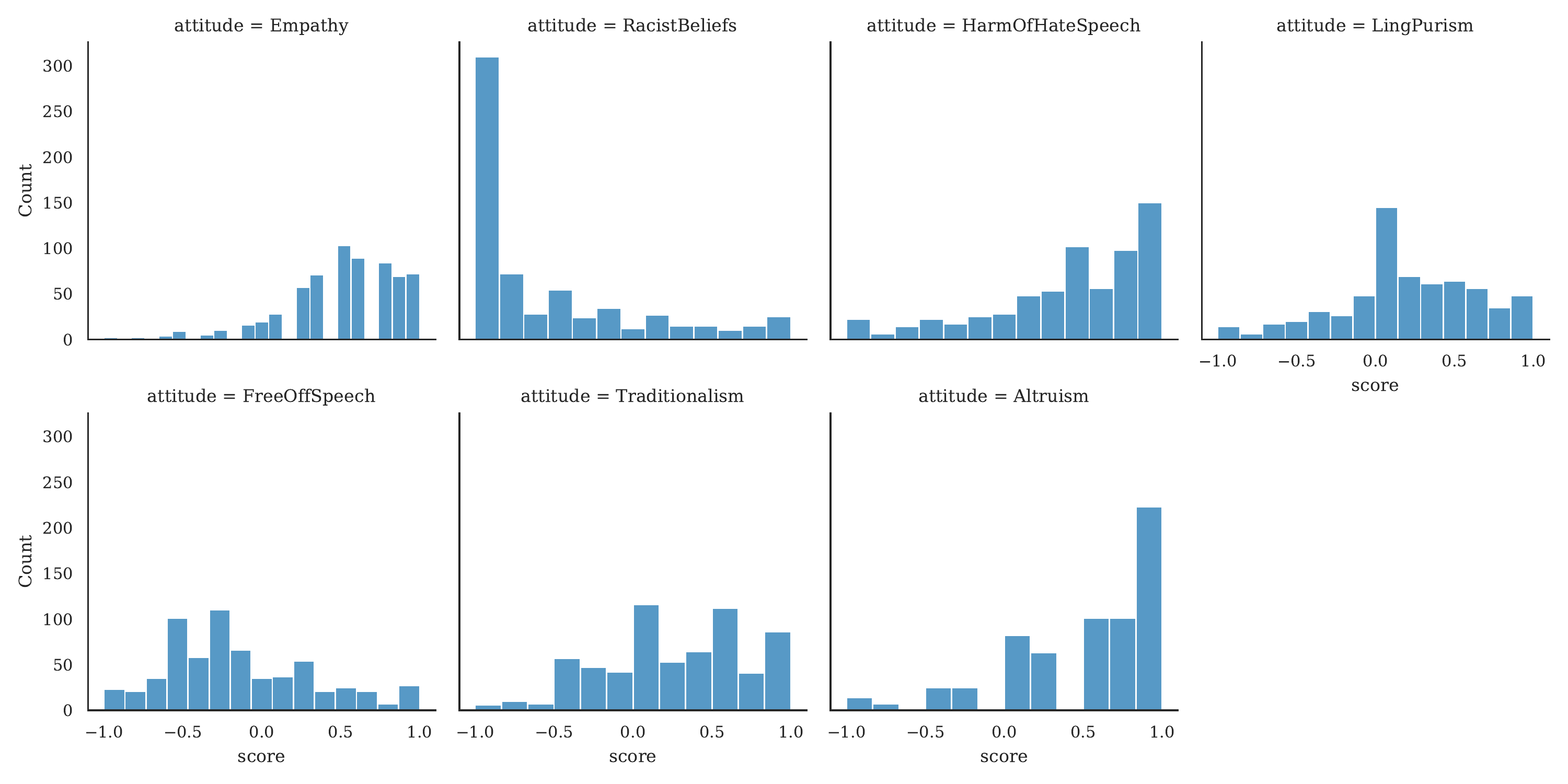}
    \caption{Distributions of the attitude scores of the workers in the \smallscale study.}
    \label{fig:smallScale-attHistograms}
\end{figure*}
\begin{table*}[t]
    \centering
    \includegraphics[width=\textwidth,trim=0cm 1.5cm 0cm 0cm,clip]{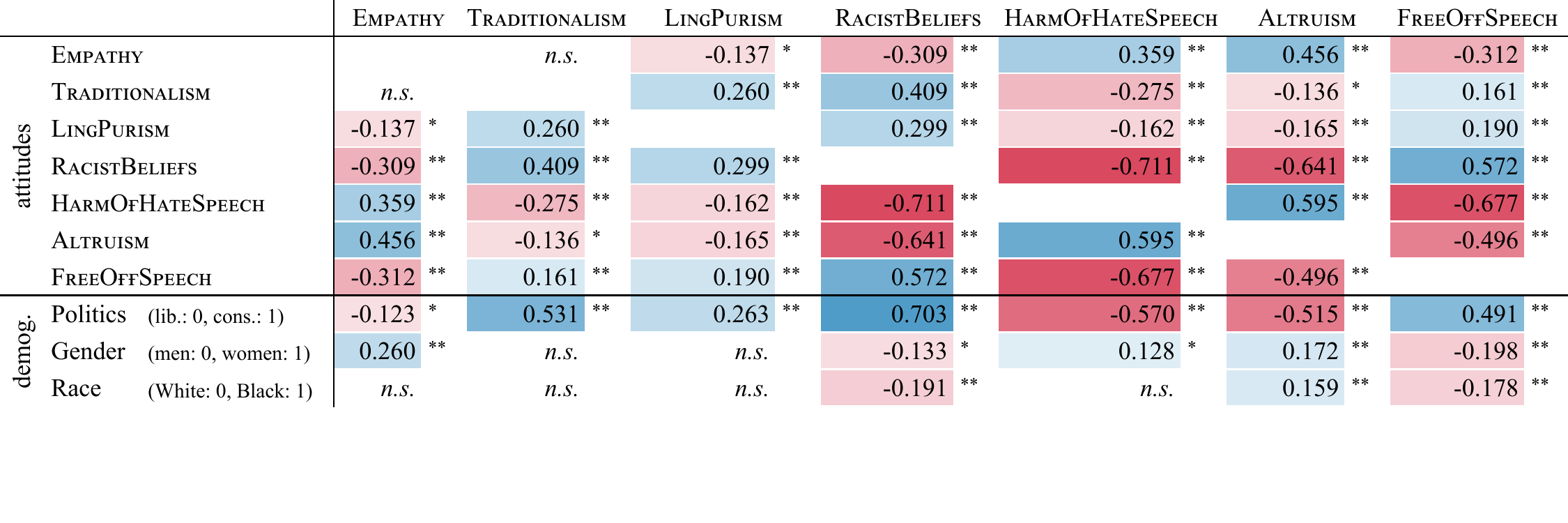}
    \caption{Pearson $r$ correlations between the attitude and demographic variables from participants in our \smallscale study.
    We only show significant correlations (*: $p$ < 0.05, **: $p$ < 0.001), and denote non-significant correlations with ``\textit{n.s.}''.
    Our demographic variables are not correlated with each other.
    }
    \label{tab:data-correls}
\end{table*}

\subsection{\empathy.} 
\label{app:empathy}
Participants were instructed to rate their level of agreement with the item statements, using a 5-point Likert scale ranging from ``\textit{not at all}'' (1) to ``\textit{very much so}'' (5).
\begin{compactitem}  
    \item Before criticizing somebody, I try to imagine how I would feel if I were in his/her place.
    \item I don't usually become sad when I see other people crying. (R)
    \item When someone is feeling `down' I can usually understand how they feel.
    \item \textbf{I have tender, concerned feelings for people or groups of people less fortunate than me.}
\end{compactitem}
This scale is an abbreviated version of the widely used Interpersonal Reactivity Index by \citet{pulos2004hierarchical}.

\subsection{\altruism.}
\label{app:altruism}
Participants were asked: ``Please tell us how important each of these is as a guiding principle in your life.''
They answered each item on a 5-point Likert scale, ranging from ``\textit{not at all important to me}'' (1) to ``\textit{extremely very important to me}'' (5).
\begin{compactitem}
    \item \textbf{Social justice, correcting injustice, caring for the weak.}
    \item Equality, equal opportunity for all.
\end{compactitem}
These items are taken from the altruism part of the scale by \cite{steg2014significance}.

\subsection{Attitude distributions \& inter-variable correlations}
\label{ssec:inter-variable-correls}

\newText{
We plot the distributions of our \smallscale participants on the seven attitude scales in Figure \ref{fig:smallScale-attHistograms}. 
While most attitudes have wider distributions, \racistBeliefs notably stands out as having a skewed distributions towards people scoring low on the scale.
}

While some attitudes may correlate highly with demographic factors at face value (e.g., \traditionalism and politically conservatism); other forms of biases may not be easily explained by demographics alone. 
We examine the relationship between our attitude measurements and annotator demographic identity variables.
Shown in Table \ref{tab:data-correls}, we find strong significant correlations between several of our annotator variables.

Notably, we find that an annotator's political orientation correlated strongly with several variables, with liberal leaning identities being associated with higher scores on the \empathy, \harmOfHateSpeech, and \altruism scales, whereas 
conservative political leaning was associated with higher scores on the \traditionalism, \lingPurism, \freeSpeech and \racistBeliefs scales.

\section{Toxicity Questions}
\label{ssec:toxicity-questions-list}

Following crowdsourcing setups in prior work \cite{waseem2016you,davidson2017automated,wulczyn2017ex,founta2018large,sap-etal-2019-risk}, we asked three fine-grained questions to annotators for each post in both our studies: 
\begin{compactitem}
    \item ``How toxic/hateful/disrespectful or offensive does this post seem \textit{to you}?''
    \item ``How much could this post be seen as toxic/hateful/disrespectful or offensive \textit{to anyone}?''
    \item ``In your opinion, how \textit{racist} is this post?''
\end{compactitem}
Given the high correlations between the two offensiveness variables (Pearson $r\geq.7$; Table~\ref{tab:offensiveness-correls}), we use create an \textit{offensiveness} (``\textit{off}.'') score by taking the average rating given to the ``to you'' and ``to anyone'' questions.
In all our analyses, we use that \textit{offensiveness} score, along with the raw \textit{racism} score.
\begin{table}[h!]
    \centering
    \includegraphics[width=\columnwidth,clip,trim=5em 6em 4.5em 5em]{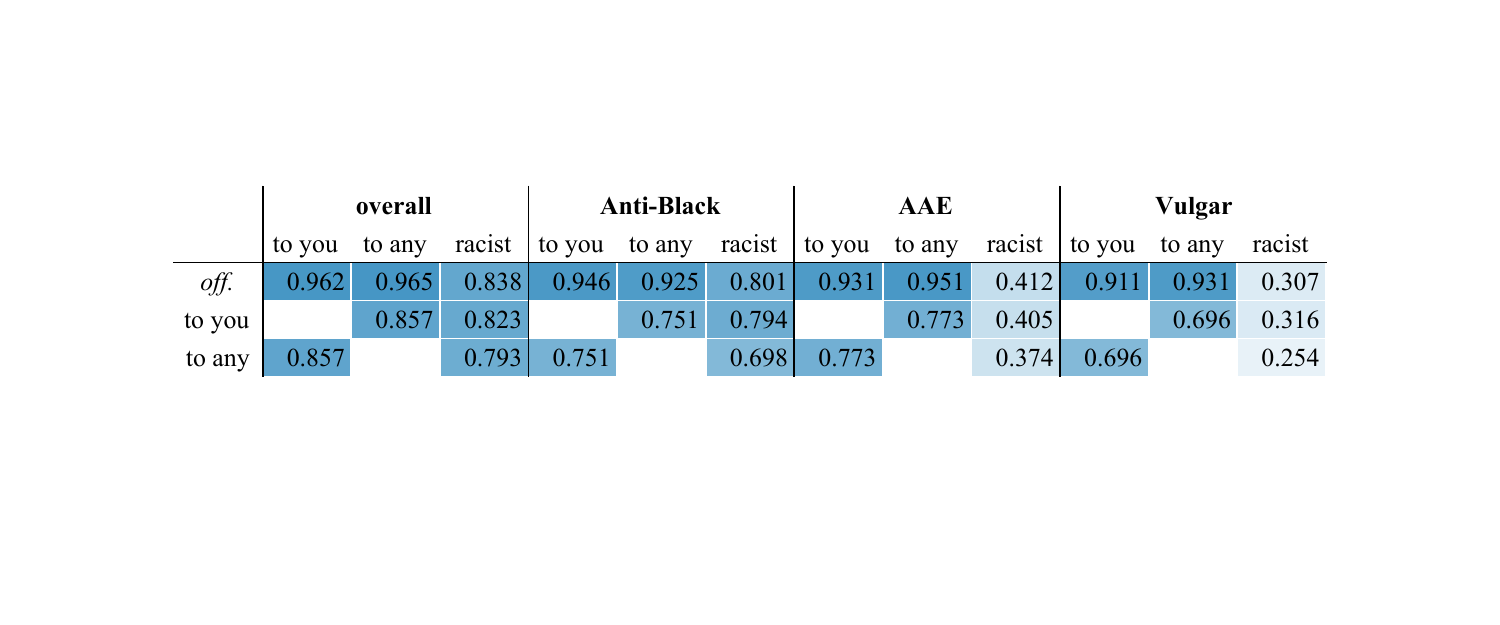}
    \caption{Correlations between different offensiveness questions for each tweet (all $p<0.001$). Since offensiveness ``\textit{to you}'' and ``\textit{to any}'' are very strongly correlated, we average them into a single offensiveness score (\textit{off.}).
    }
    \label{tab:offensiveness-correls}
\end{table}

\section{Small-Scale Controlled Study Details}
\label{sec:smallScale-moreSetup}

\subsection{Data Selection \& Validation}
\label{ssec:smallScale-data-selection}

We aimed to select online posts that were very indicative of each of the above characteristics (\vulgar, \AAE, \racist) but not indicative of the others, in order to tease out the effect of that category.
We selected \vulgar and \AAE posts from a publicly available large corpus of tweets annotated for hate speech by \citet{founta2018large}.\footnote{We only used the training subset of the corpus.}
For each tweet in that corpus, we detected the presence of non-identity related profanity or swearwords using the list from \citet{zhou2021challenges}, and extracted the likelihood that the tweet is in \AAE using a lexical detector by \citet{blodgett2016demographic}.
As candidates, we selected 10 \vulgar tweets that have low likelihood of being \AAE, and 26 tweets that have high likelihood of being \AAE but contain no vulgarity.
For \racist posts, we selected 11 candidate online posts curated by \citet{Zevallos2017sociologyOfRace}. 

We ran a human validation study to verify that the candidate posts are truly indicative of their respective categories.
We created an annotation scheme to collect binary ratings for two questions per post: ``does it contain vulgar language'', and ``is it offensive to minorities''; a post could belong to either category or neither.
Each post was manually annotated by three undergraduate research assistants trained for the task.
Post validation, we manually selected 5 posts per category with perfect inter-annotator agreement.
Table~\ref{tab:study-tweets} lists the final 15 posts used for our study.

\subsection{Participant Recruitment}
We ran our study on Amazon Mechanical Turk (MTurk), a crowdsourcing platform that is often used to collect offensiveness annotations.\footnote{Note, this study was approved by the author's institutional review board (IRB).}
With the task at hand, we sought a racially and politically diverse pool of participants, which can be challenging given that MTurk workers are usually tend to be predominantly white and skew liberal \cite{huff2015these,burnham2018mturk,loepp2020distinction}.
Therefore, we ran a pre-selection survey to collect race and political ideology of workers, noting that this pre-survey could grant them access to a longer survey on free speech, hate speech, and offensiveness in language.\footnote{To better recruit for our pre-survey, we noted in the title that ``BIPOC people and conservatives were encouraged to participate,'' and also varied the title's wording to emphasize free speech or hate speech in different recruiting rounds.}$^,$\footnote{We compensated workers \$0.02--\$0.03 for this pre-survey.}
We stopped recruiting once we reached at least 200 Black and 200 conservative participants.

\subsection{Study Setup}
We ran our study on the widely used survey platform Qualtrics, using an MTurk HIT to recruit and compensate participants.\footnote{Participants were compensated \$4.33 for the entire survey, equivalent to an average hourly compensation of \$22/h.}
Participants were first asked to consent to the task, then were shown instructions for annotating the 15 posts (with occasional reminders of the instructions).
Then we asked participants their views on several topics using the scales described in \S\ref{ssec:why} and \S\ref{app:attitude_scales} and finally their demographics. 
Throughout the study, we added three attention checks to ensure the quality of responses.

Allowing only Black, white liberal, and white conservative workers to participate, we ran our survey for 4 weeks from March 10 to April 5, 2021, occasionally reminding participants from our pre-survey that they could take the survey.


\section{\Largescale Annotation Study Details}
\label{sec:largeScale-moreSetup}

\subsection{Data Selection}
\label{ssec:largeScale-dataSelection}
In this study, we draw from two existing corpora of posts labeled for toxicity, hate, or offensiveness.
First, we select posts that are automatically detected as \AAE and/or \vulgar from \citet{founta2018large}, using the lexical detector of \AAE by \cite{blodgett2016demographic} and the \vulgar wordlist by \citet{zhou2021challenges}.
Second, we select posts that are automatically detected as \vulgar and/or annotated as \racist from \citet{vidgen-etal-2021-learning}.
Importantly, in this large-scale study, we consider posts that potentially have multiple characteristics (e.g., \AAE and \vulgar), and thus consider both posts with potentially offensive identity references (\vulgar-\OI) as well as non-identity vulgar words (\vulgar-\OnI). 
However, to circumvent potential racial biases in what is labelled as ``racist'' in the \citet{vidgen-etal-2021-learning} corpus \cite{sap-etal-2019-risk,davidson-etal-2019-racial}, we do not consider posts that are annotated as \racist but detected as \AAE.

Given an initial set of posts from our categories, we then randomly sample up to 600 posts, stratifying by toxicity label, vulgarity, \AAE, and \racist meaning.
Our final sample contains 571 posts, as outlined in Table \ref{tab:mturk-counts} and Figure \ref{fig:mturk-cat-counts}.

\subsection{\Largescale Survey Details}
As in the \smallscale study, we recruit participants using a pre-qualifying survey on MTurk. 
Then, we set up a second MTurk task to collect toxicity ratings, and annotator attitudes and identities.
For each post, we collected two ratings from white conservative workers, two from white liberal workers, and two from Black workers.
To better mirror the crowdsourcing setting and to reduce the annotator burden, we shorten the task to only ask one question per attitude (listed in  \S\ref{app:attitude_scales}). 
We also asked one attention check question to ensure data quality.

For this study, our final dataset contains 3,171 ratings from $N=173$ participants.\footnote{As before, we discard 255 ratings where workers failed an attention check.}
Our participants were 53\% were men, 45\% women, and $<$2\% non-binary, identified as 76\% white, 20\% Black, and $<$4\% some other race, and spanned the political spectrum from 54\% liberal to 30\% conservative, with 16\% centrists or moderates.

\subsection{Selecting Attitude Questions}
\label{ssec:largeScale-itemSelection}
In order to simplify the annotation task for annotators, we abridged the attitude scales to only one item.
Using the data from the \smallscale study, we select the question that best correlated with all toxicity ratings.
Specifically, for each scale, we first take the tweet category with the highest correlation with toxicity (e.g., \racist posts for \racistBeliefs), and then take the item whose response scores correlated most with the toxicity rating for those posts.
Those items are bolded in \S\ref{app:attitude_scales}.

\section{Further \Smallscale Results}
\label{sec:smallScale-more}

We show all associations between attitudes and toxicity ratings in Table~\ref{tab:smallScale-full}.

\begin{table*}[t]
    \centering
    \includegraphics[width=\textwidth,clip,trim=.7em 0em .7em 0em]{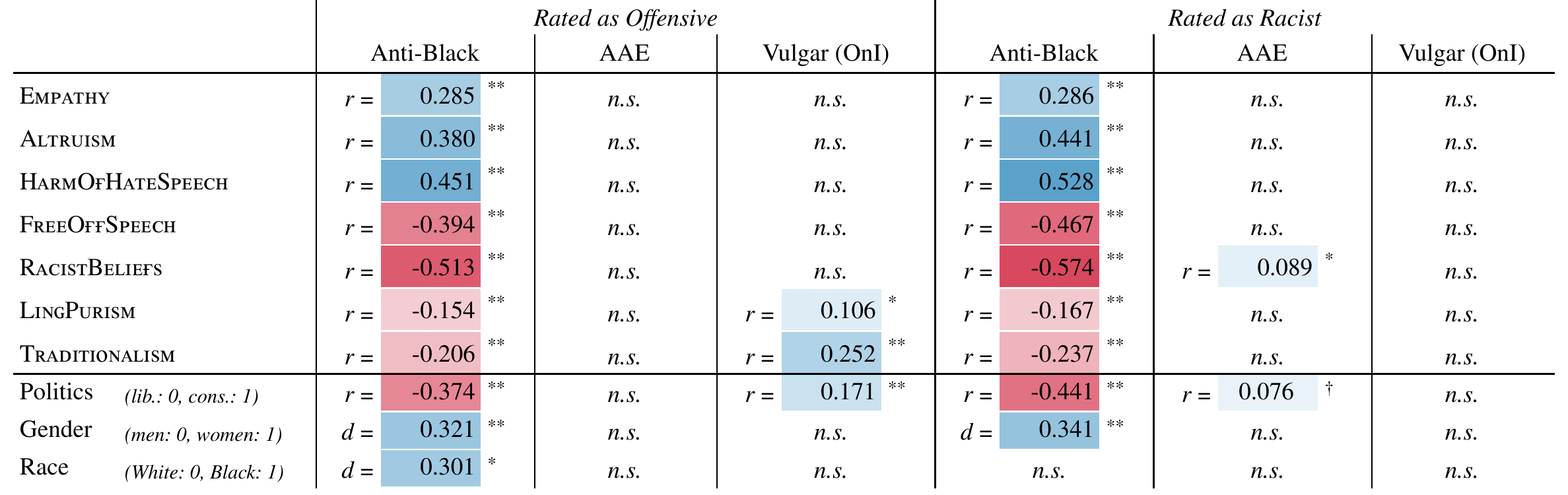}
    \caption{Full set of results from our analyses of the \smallscale study of 15 posts, presented as Pearson $r$ or Cohen's $d$ effect sizes, along with significance levels ($^\dagger$: $p<0.075$, $^*$: $p<0.05$, $^{**}$: $p<0.001$).
    We correct for multiple comparison for variable relationships that were exploratory (i.e., not discussed as hypotheses in \S\ref{sec:racist-results}--\ref{sec:vulgar-results}).
    }
    \label{tab:smallScale-full}
\end{table*}

\begin{figure}[t]
    \centering
    \includegraphics[width=\columnwidth,clip,trim=3em 3em 3em 3em]{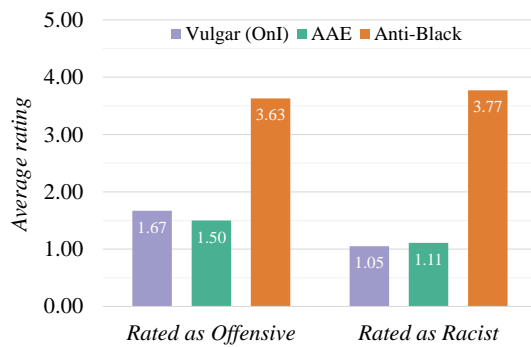}
    \caption{Average ratings of offensiveness and racism for each tweet category in the \smallscale controlled study.
    All differences are significant ($p<0.001$) after correcting for multiple comparisons.
    }
    \label{fig:smallScale-offAvgs}
\end{figure}

Additionally, we investigate the differences in the overall toxicity ratings of \racist vs. \AAE vs. \vulgar posts? (Figure~\ref{fig:smallScale-offAvgs}).
Overall, \racist tweets were rated as substantially more offensive and racist than \AAE or \vulgar tweets (with effect sizes ranging from $d$ = 2.4 to $d$ = 3.6).
Additionally, \vulgar tweets were rated as more offensive than \AAE tweets ($d$ = -0.29, $p$ < 0.001).

Surprisingly, we also found that \AAE tweets were considered slightly more racist than \vulgar tweets ($d$ = 0.19, $p$ < 0.001).
To further inspect this phenomenon, we performed exploratory analyses by computing the differences in ratings of racism for \AAE and \vulgar broken down by annotator gender, race, and political leaning. 
We found that \AAE tweets were rated as significantly more racist than \vulgar tweets only by annotators who were white or liberal ($d$ = 0.20 and $d$ = 0.22, respectively, with $p$ < 0.001 corrected for multiple comparisons), compared to Black or conservative.
There were no significant differences when looking at men and women separately.


\section{Further \Largescale Results}
\label{app:largeScale-results}

To account for the varying number of posts that each annotators could rate, we use a linear mixed effects model\footnote{Using the Python \href{https://www.statsmodels.org/devel/generated/statsmodels.regression.mixed_linear_model.MixedLM.html}{statsmodels implementation}.} 
to compute associations between each post's toxicity ratings and identities or attitudes.
Specifically, we our linear model regresses the attitude score onto the toxicity score, with a random effect for each worker.\footnote{In R-like notation, \texttt{toxicity$\sim$attitude+(1|WorkerId)}.}

\begin{figure}[t]
    \centering
    \includegraphics[width=\columnwidth,page=2]{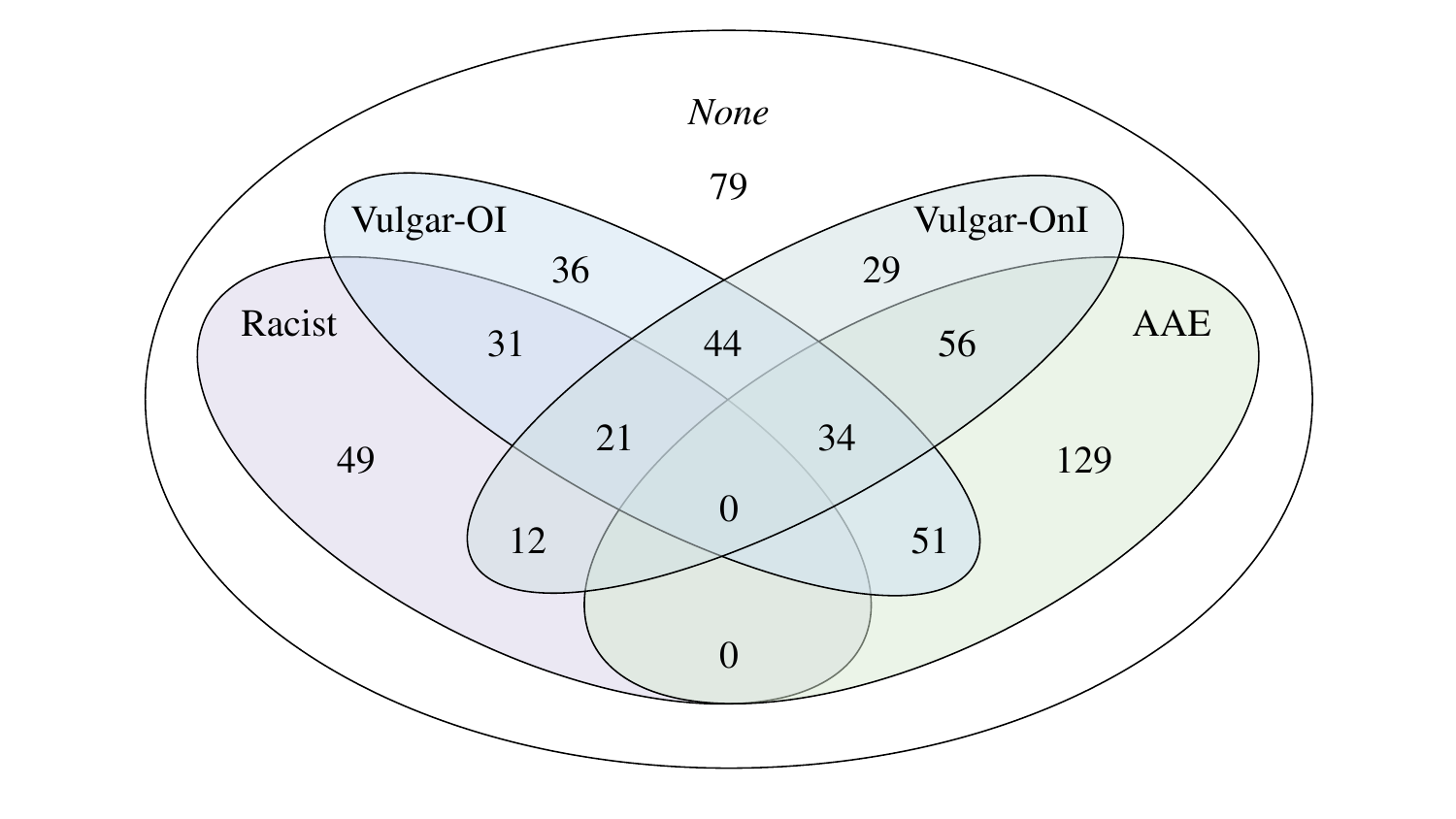}
    \caption{Venn diagram of number of tweets in each of the categories.}
    \label{fig:mturk-cat-counts}
\end{figure}

See Figure~\ref{fig:mturk-cat-counts} and Table~\ref{tab:largeScale-full}.

\begin{table*}[t]
    \centering
    \includegraphics[width=\textwidth,clip,trim=0em 0em 0em 0]{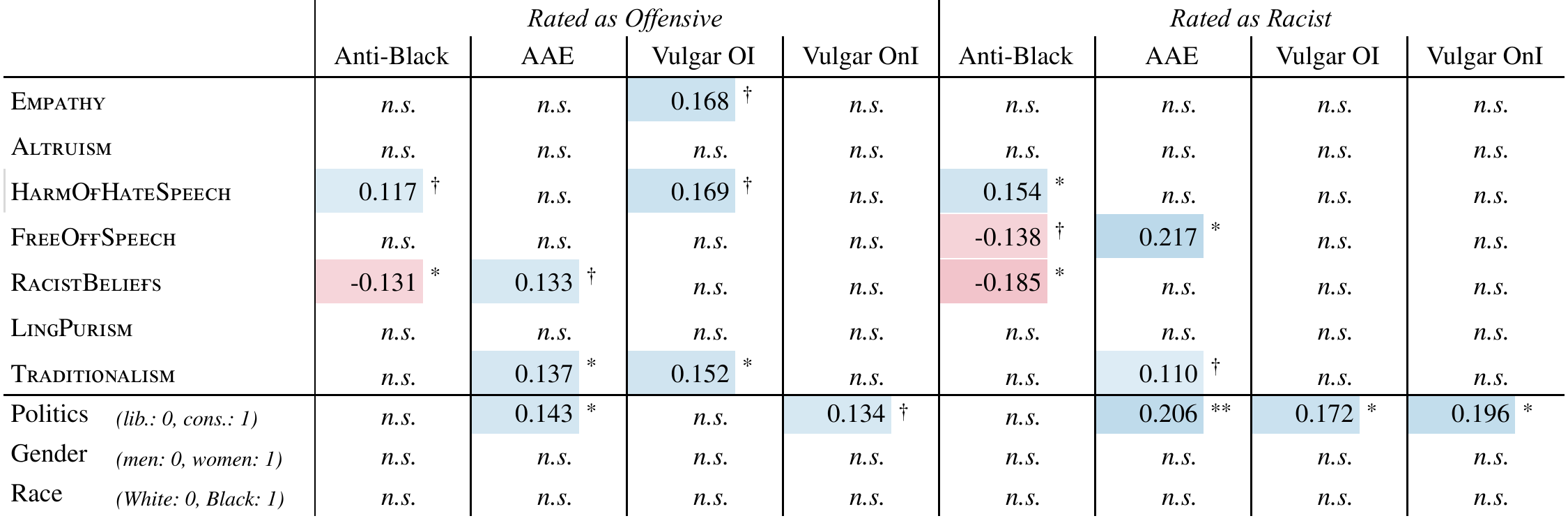}
    \caption{
    Associations between the annotator demographic and attitude variables and their ratings of offensiveness and racism on the posts from the \largescale study. We break down the results by category, but categories are overlapping.
    Only significant associations ($\beta$ coefficients from a mixed effects model) are shown ($^\dagger$: $p<0.075$, $^*$: $p<0.05$, $^{**}$: $p<0.001$; Holm-corrected for multiple comparisons).
    }
    \label{tab:largeScale-full}
\end{table*}




\section{\perspectiveAPI Case Study: Details \& Results}
\label{sec:perspective-details-results}
\subsection{Details}
\label{sec:perspective-details}

We first obtain \perspective toxicity scores for all the posts in our \largescale study (\S\ref{ssec:large-scale-design}).\footnote{the API was accessed in October 2021}
Then, we split workers into two different groups for each of our attitudes and identity dimensions.
For attitudes and political leaning, we assign each annotator to a ``high'' or ``low'' group based on whether they scored higher or lower than the mean score on that attitude scale.
For gender and race, we use binary bins for man/woman and white/black.

Then, for each attitude or identity dimension, we compute the Pearson $r$ correlation between the \perspective score and the toxicity ratings from the high and low groups, considering posts from potentially overlapping categories (e.g., \AAE and potentially \vulgar posts).
Finally, we compare the high and low correlations using Fisher's $r$-to-$z$ transformation \cite{silver1987averaging}.

\subsection{Results}
\label{sec:perspective-results}
\begin{figure*}[ht]
    \centering
    \includegraphics[width=\textwidth]{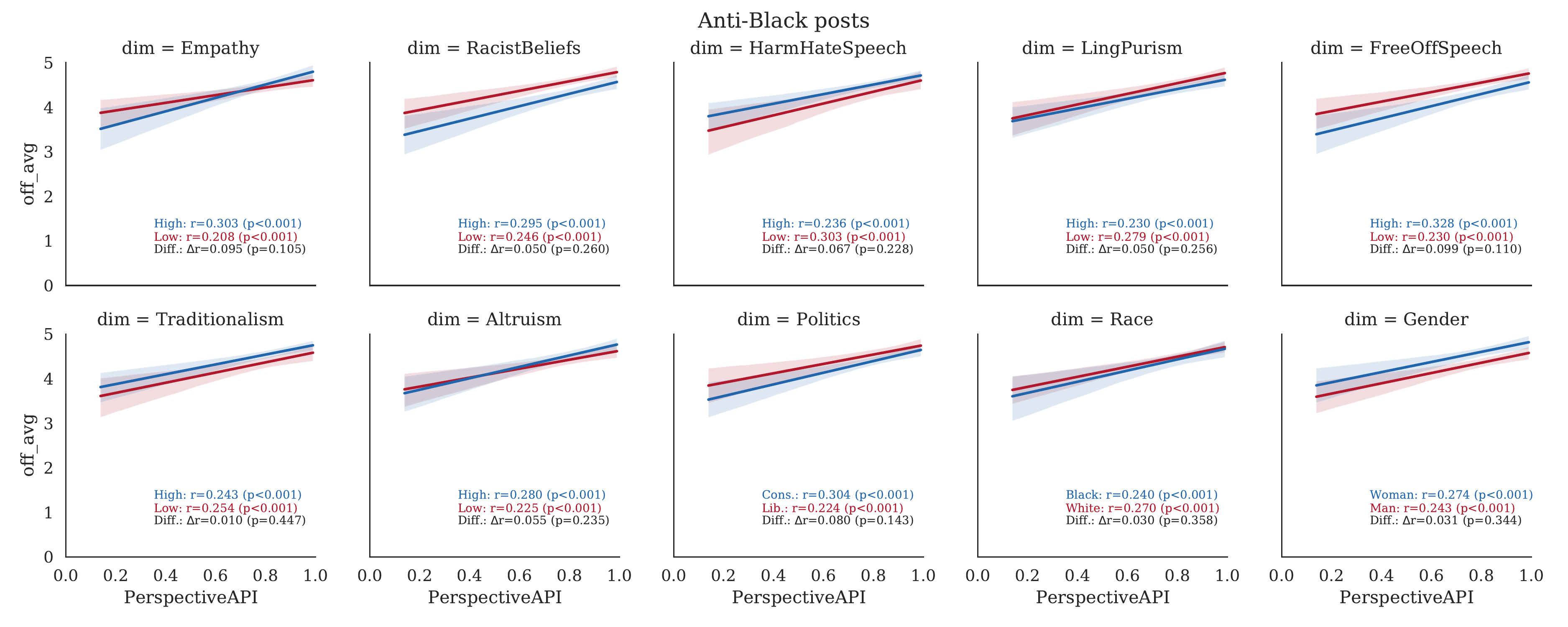}
    \caption{\perspectiveAPI and ratings of \textit{offensiveness} of \racist tweets.}
    \label{fig:persp-racist-off}
\end{figure*}

\begin{figure*}[ht]
    \centering
    \includegraphics[width=\textwidth]{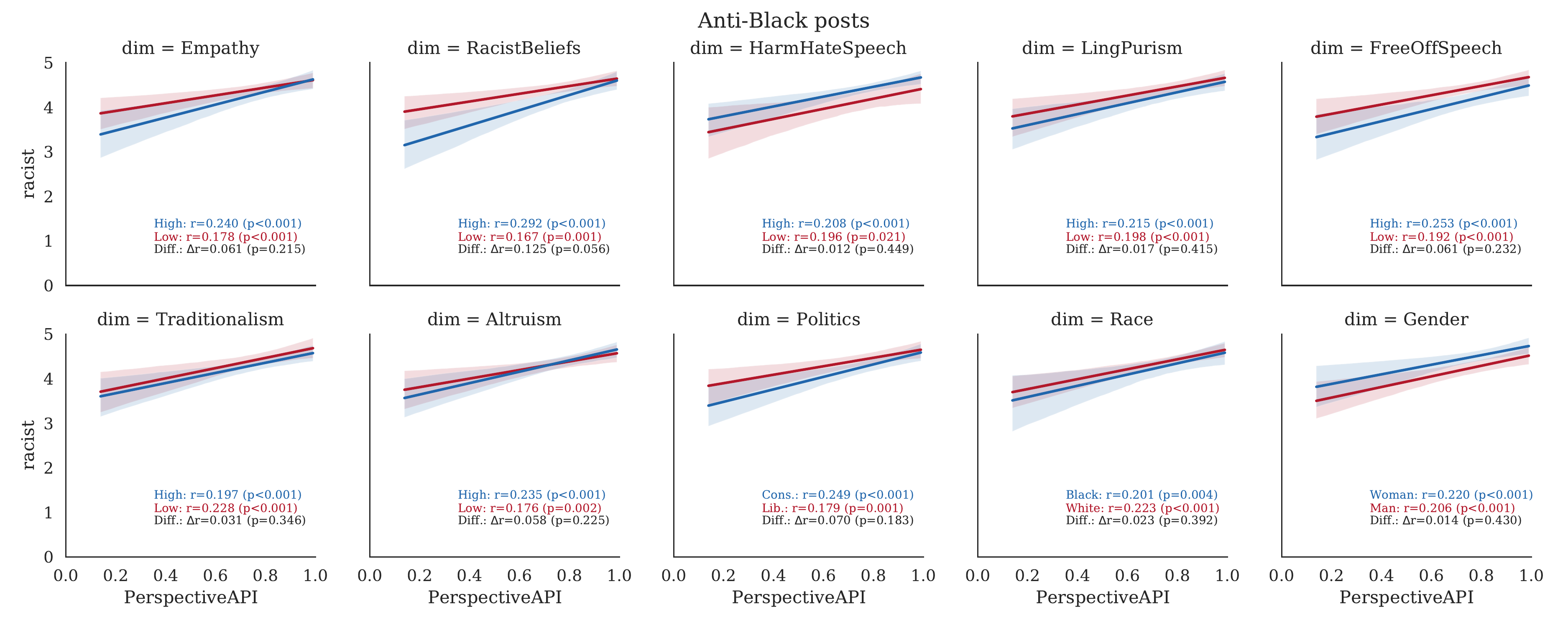}
    \caption{\perspectiveAPI and ratings of \textit{racist} of \racist tweets.}
    \label{fig:persp-racist-racism}
\end{figure*}

\begin{figure*}[ht]
    \centering
    \includegraphics[width=\textwidth]{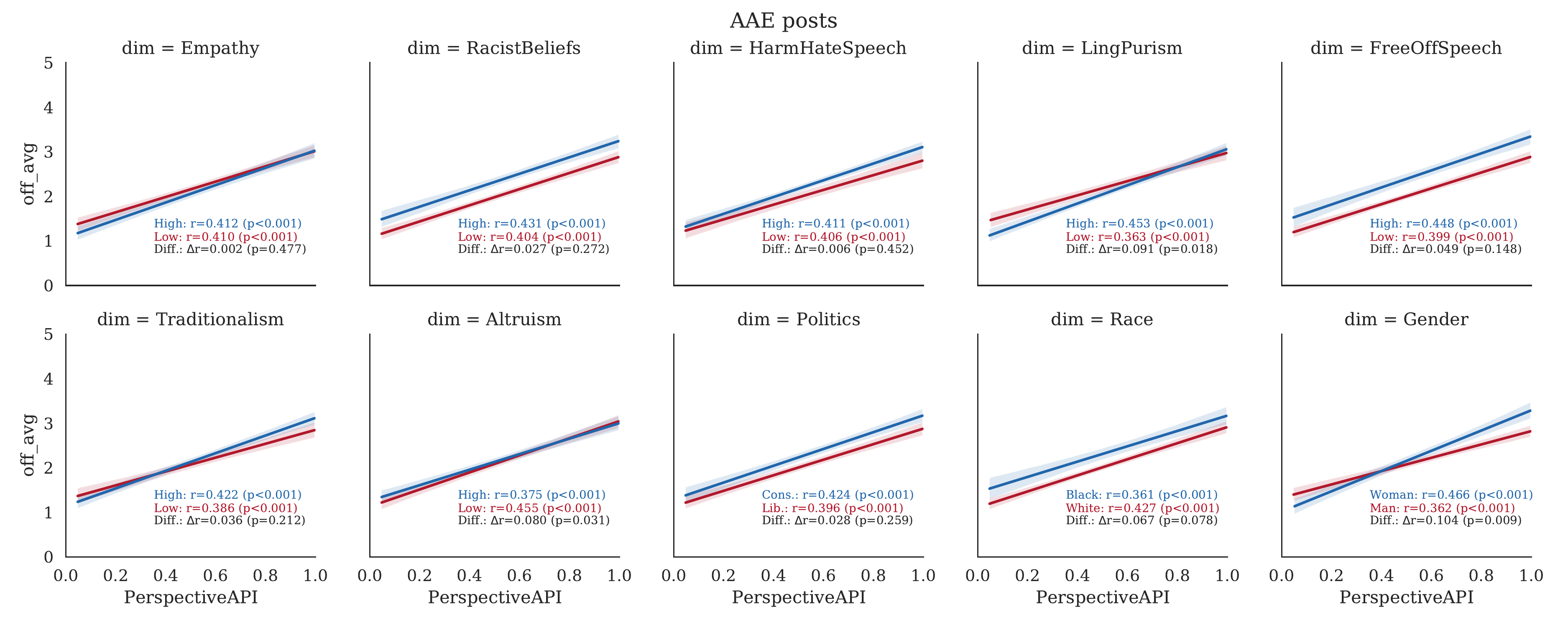}
    \caption{\perspectiveAPI and ratings of \textit{offensiveness} of \AAE tweets.}
    \label{fig:persp-aae-off}
\end{figure*}

\begin{figure*}[ht]
    \centering
    \includegraphics[width=\textwidth]{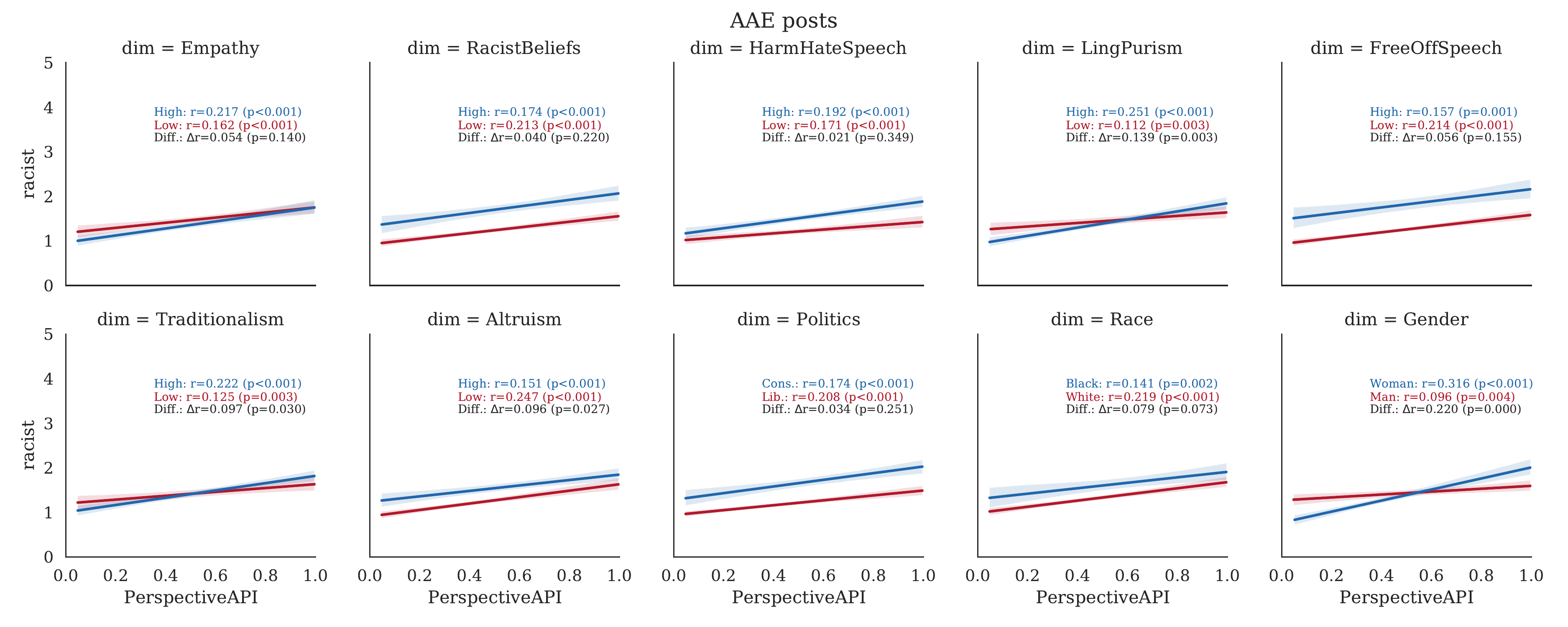}
    \caption{\perspectiveAPI and ratings of \textit{racist} of \AAE tweets.}
    \label{fig:persp-aae-racism}
\end{figure*}

\begin{figure*}[ht]
    \centering
    \includegraphics[width=\textwidth]{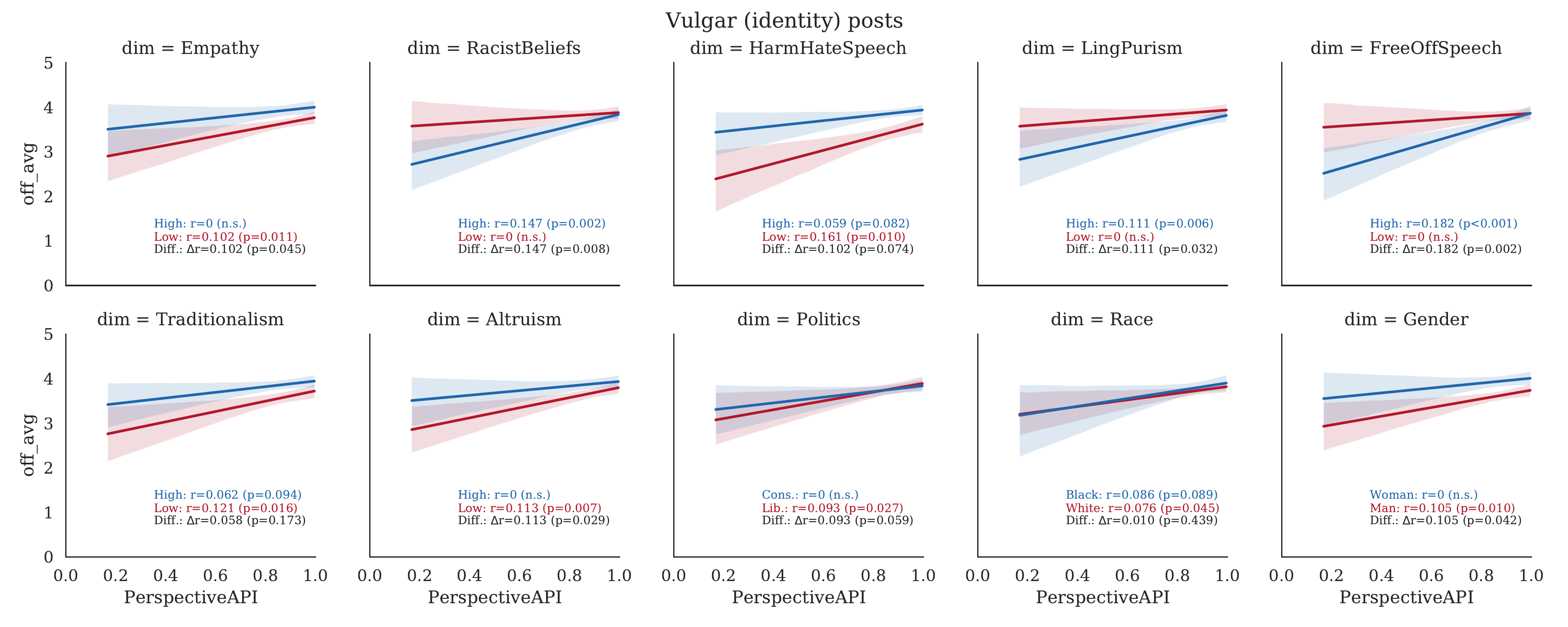}
    \caption{\perspectiveAPI and ratings of \textit{offensiveness} of \vulgar-\OI tweets.}
    \label{fig:persp-vulgaroi-off}
\end{figure*}

\begin{figure*}[ht]
    \centering
    \includegraphics[width=\textwidth]{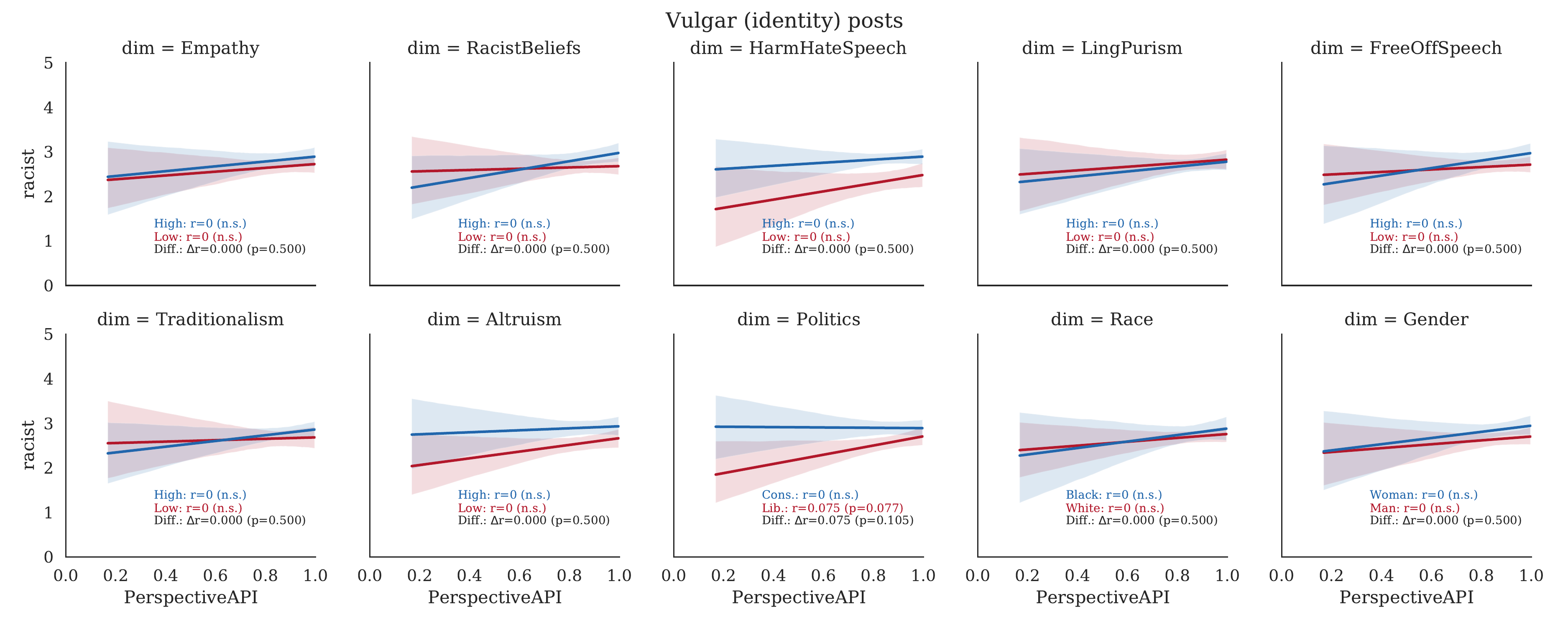}
    \caption{\perspectiveAPI and ratings of \textit{racist} of \vulgar-\OI tweets.}
    \label{fig:persp-vulgaroi-racism}
\end{figure*}

\begin{figure*}[ht]
    \centering
    \includegraphics[width=\textwidth]{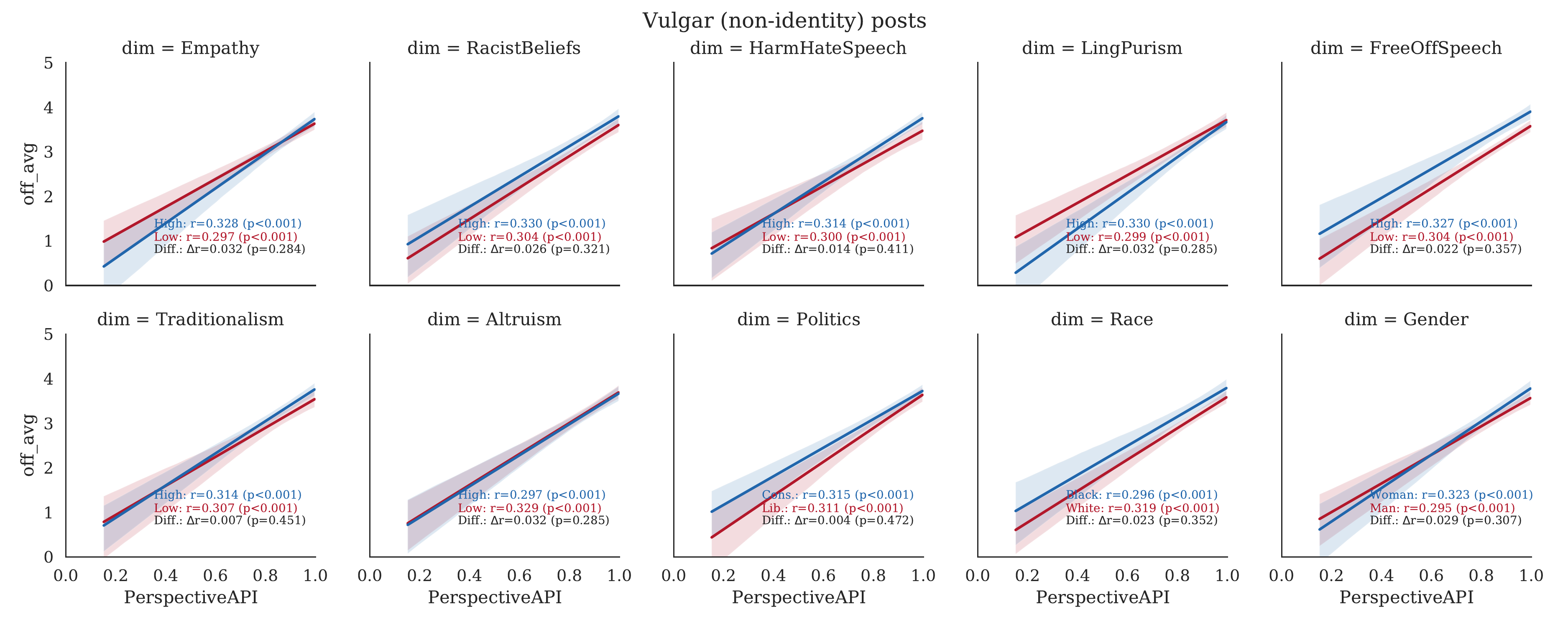}
    \caption{\perspectiveAPI and ratings of \textit{offensiveness} of \vulgar-\OnI tweets.}
    \label{fig:persp-vulgaroni-off}
\end{figure*}

\begin{figure*}[ht]
    \centering
    \includegraphics[width=\textwidth]{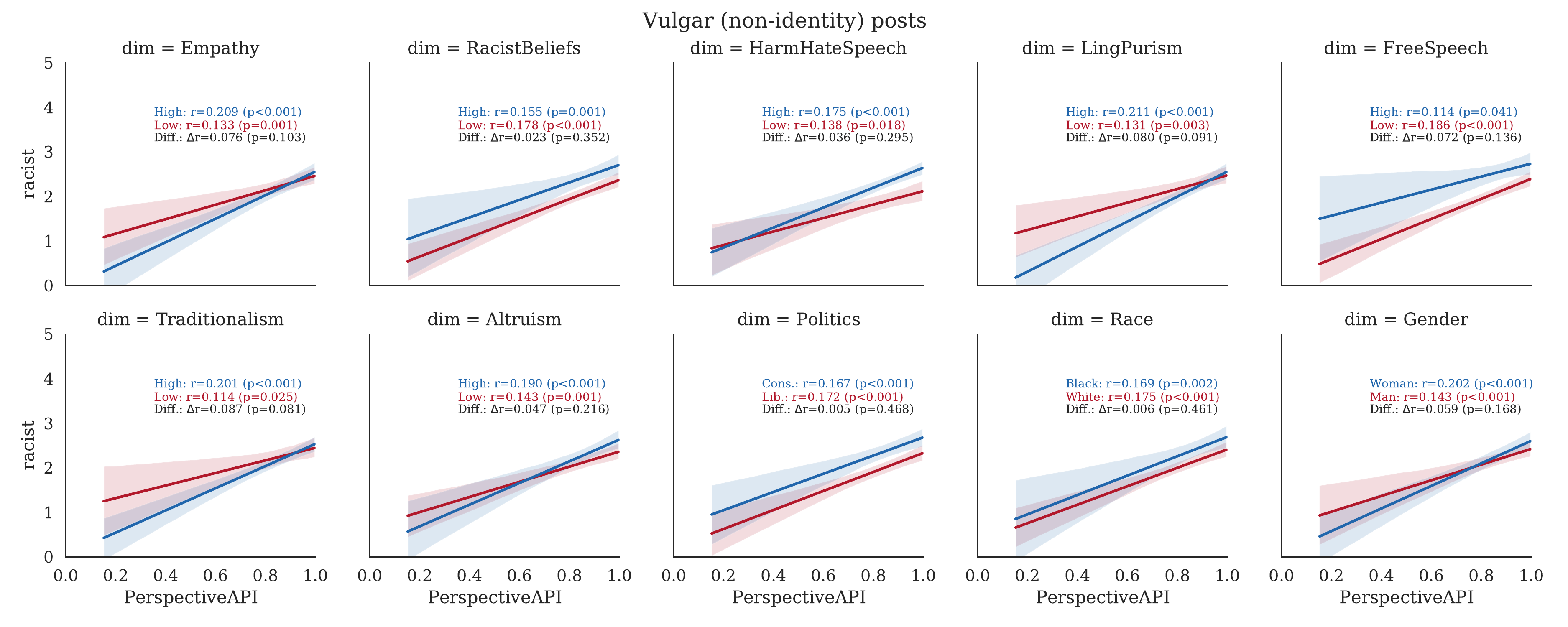}
    \caption{\perspectiveAPI and ratings of \textit{racist} of \vulgar-\OnI tweets.}
    \label{fig:persp-vulgaroni-racism}
\end{figure*}

See Table~\ref{tab:perspectiveCorrelDiffs} and Figures \ref{fig:persp-racist-off}--\ref{fig:persp-vulgaroni-racism}.

\begin{table*}[t]
    \centering
    \includegraphics[width=\textwidth]{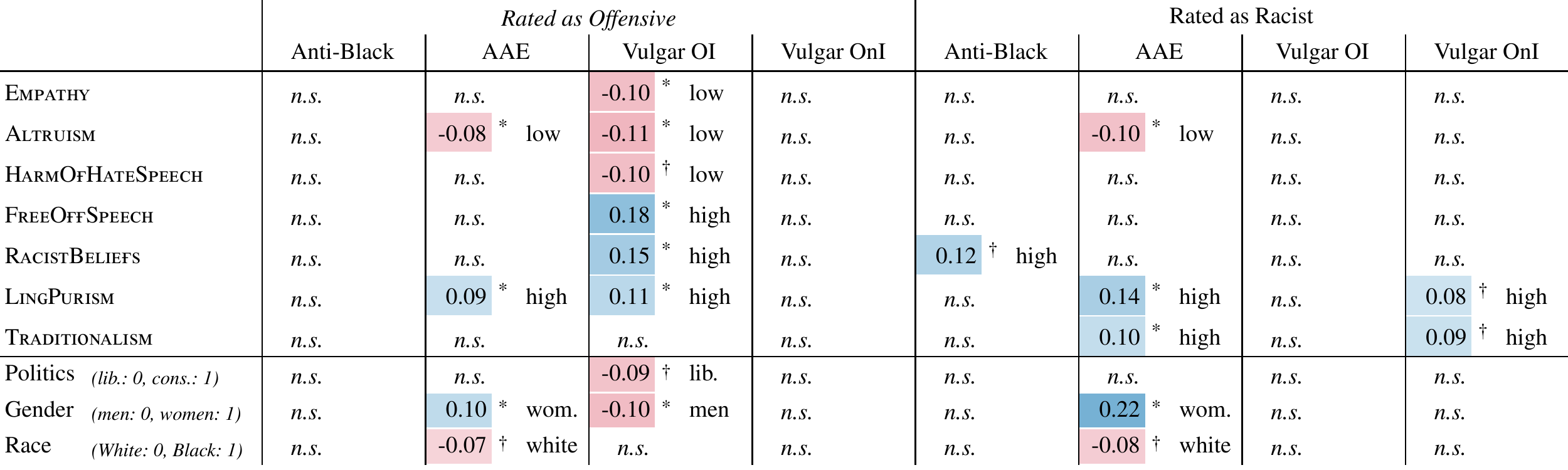}
    \caption{We correlated the \perspectiveAPI toxicity scores with offensiveness/racism ratings by our annotators, breaking them into two bins based on their attitude scores. Then, we used Fisher's z-to-r test to measure whether the differences in correlations between the annotators who are high/low were significant ($^\dagger$: $p<0.1$, $^*$: $p<0.05$).
    }
    \label{tab:perspectiveCorrelDiffs}
\end{table*}

\end{document}